\documentclass{article} 
\usepackage{iclr2025_conference,times}

\usepackage{amsmath,amsfonts,bm}

\def\eqref#1{equation~\ref{#1}}

\def\1{\bm{1}}

\DeclareMathAlphabet{\mathsfit}{\encodingdefault}{\sfdefault}{m}{sl}
\SetMathAlphabet{\mathsfit}{bold}{\encodingdefault}{\sfdefault}{bx}{n}

\usepackage{hyperref}
\usepackage{url}
\usepackage{subcaption}

\usepackage{graphicx}
\usepackage{booktabs}
\usepackage{multirow}
\usepackage{amsmath}
\usepackage{enumitem}

\newcommand{\internlogo}{
  \raisebox{-0.3cm}{\includegraphics[width=1.3cm,height=1.2cm]{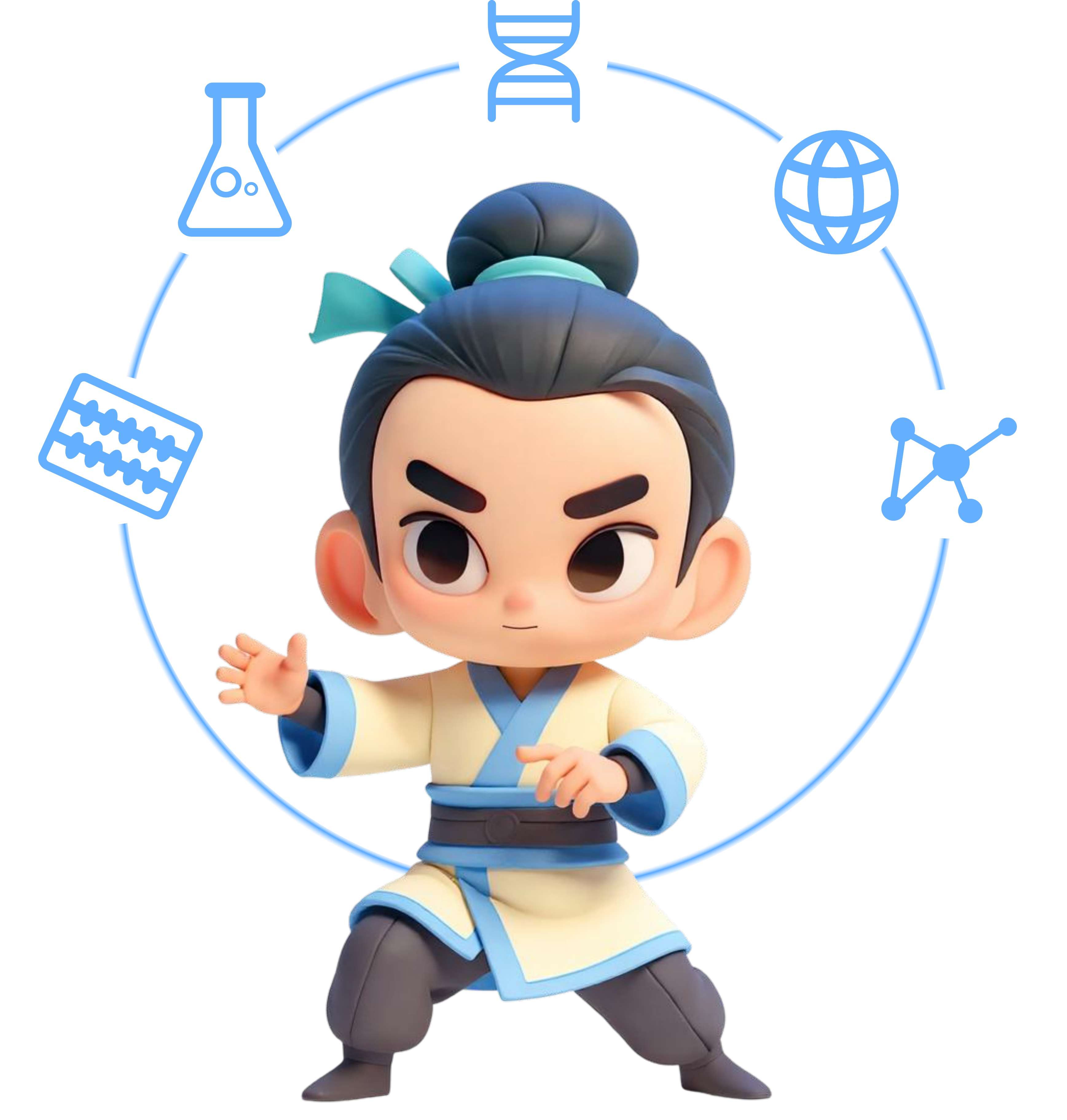}}
}

\title{
  \begin{tabular}{@{}l@{\hspace{0.5cm}}l@{}}
    \multirow{2}{*}{\internlogo} & 
    Intern-S1: A Scientific \\
    &  Multimodal Foundation Model
  \end{tabular}
}

\author{Intern-S1 Team, Shanghai AI Laboratory}

\iclrfinalcopy 
\begin{document}

\maketitle

\begin{abstract}
In recent years, a plethora of open-source foundation models have emerged, achieving remarkable progress in some widely attended fields, with performance being quite close to that of closed-source models.
However, in high-value but more challenging scientific professional fields, either the fields still rely on expert models, or the progress of general foundation models lags significantly compared to those in popular areas, far from sufficient for transforming scientific research and leaving substantial gap between open-source models and closed-source models in these scientific domains.
To mitigate this gap and explore a step further toward Artificial General Intelligence (AGI), we introduce Intern-S1, a specialized generalist equipped with general understanding and reasoning capabilities with expertise to analyze multiple science modal data.
Intern-S1 is a multimodal Mixture-of-Experts (MoE) model with 28 billion activated parameters and 241 billion total parameters, continually pre-trained on 5T tokens, including over 2.5T tokens from scientific domains. 
In the post-training stage, Intern-S1 undergoes offline and then online reinforcement learning (RL) in InternBootCamp, where we propose Mixture-of-Rewards (MoR) to synergize the RL training on more than 1000 tasks simultaneously. Through integrated innovations in algorithms, data, and training systems, Intern-S1 achieved top-tier performance in online RL training.
On comprehensive evaluation benchmarks, Intern-S1 demonstrates competitive performance on general reasoning tasks among open-source models and significantly outperforms open-source models in scientific domains, surpassing closed-source state-of-the-art models in professional tasks, such as molecular synthesis planning, reaction condition prediction, predicting thermodynamic stabilities for crystals. Our models are available at \url{https://huggingface.co/internlm/Intern-S1}. 

\end{abstract}

\section{Introduction}

Scientific research, recognized as one of the ultimate goals in the development of artificial general intelligence (AGI) due to its potential to drive fundamental breakthroughs in human society, imposes uniquely stringent demands on AI systems. It requires models not only to understand and capture the intrinsic laws underlying diverse but low-resource distributed scientific modalities—ranging from molecular structures to time-series signals—but also perform long-term, rigorous reasoning processes, such as hypothesis validation and experimental design optimization. These requirements collectively necessitate the development of a \textbf{multimodal large reasoning model} capable of comprehending scientific modalities, serving as a foundational tool to accelerate scientific discovery.

Over the past few years, open-source multimodal large models, primarily centered on vision-language modalities, and large reasoning models (LRMs) have achieved rapid progress. Notably, in areas of widespread public attention—such as natural image understanding, mathematical problem-solving, and code generation—these open-source models~\citep{zhu2025internvl3, guo2025deepseek, yang2025qwen3} have approached or even partially surpassed their closed-source counterparts~\citep{openai2024gpt4technicalreport, o3, claude4, gemini2.5}. This advancement has sparked growing expectations for their application in the more challenging science domains. However, in high-value yet more challenging scientific scenarios, the progress of open-source foundation models lags significantly behind their development in popular domains such as mathematics and code. Moreover, a substantial gap remains between open-source and closed-source models~\citep{o3,grok4,gemini2.5} in these scientific areas, limiting the former to contribute meaningfully to cutting-edge research.

\begin{figure}[htb]
    \centering
    \includegraphics[width=1.0\linewidth]{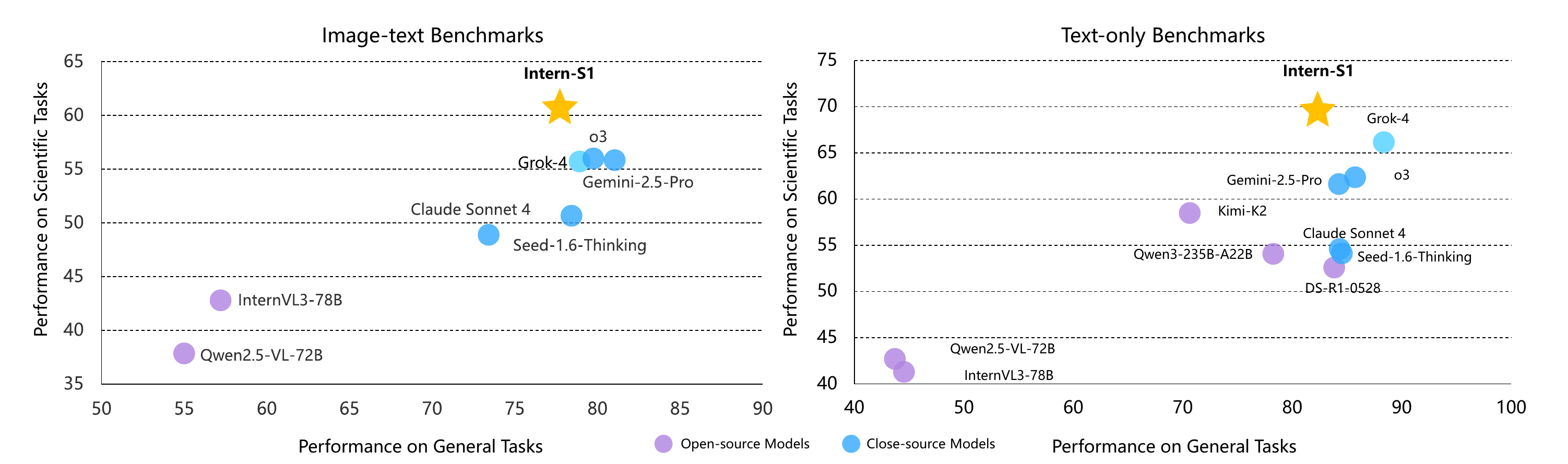}
    \caption{Performance comparison among open-source and close-source models on Image-text and Text-only Benchmarks. Results demonstrate that Intern-S1 has a top-tier general reasoning capability among open-source models and outperforms closed-source models in scientific domains. General benchmarks: MMLU-Pro (text-only), GPQA (text-only), AIME2025 (text-only), MMMU, MMStar
Science benchmarks: SmolInstruct (text-only), ChemBech (text-only), MatBench (text-only), SFE, Physics}
    \label{fig:intro_main_results}
\end{figure}

To bridge this gap in scientific understanding and reasoning capabilities between open-source and closed-source models, and to propel open-source models one step closer to AGI, in this report, we share our experience and key findings from building Intern-S1, an open-source scientific multi-modal model designed for solving complex scientific tasks, which is capable of processing images, text, and scientific data, including non-natural visual data, molecular structures, and time-series signals. As shown in the Figure~\ref{fig:intro_main_results}, Intern-S1 outperforms both open-source and close-source models on image-text or text-only scientific tasks. 

\begin{figure}[h]
    \centering
    \includegraphics[width=0.8\linewidth]{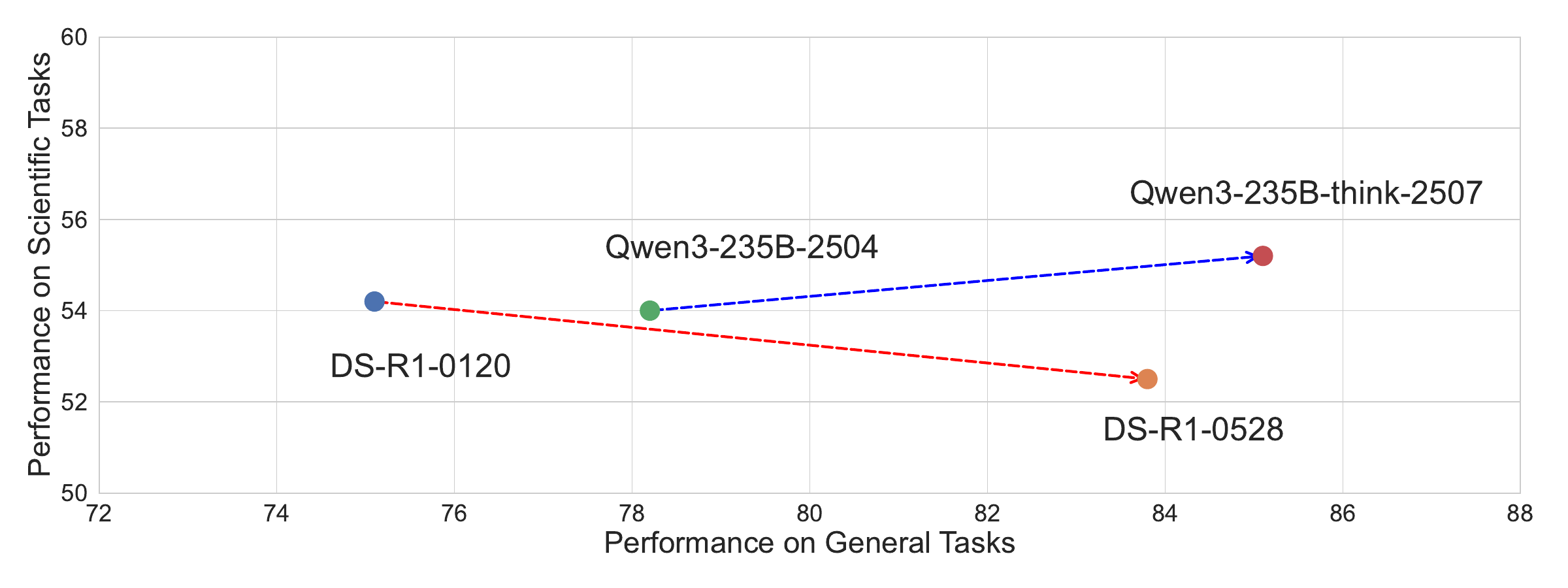}
    \caption{Performance trend of LLMs across popular and low-resource (science) tasks. The X-axis is the average of three popular general benchmarks, MMLU-Pro, GPQA,  AIME2025. The Y-axis is the average of three benchmarks in science domain, SmolInstruct, ChemBench, MatBench. Although the top-tier open-source LLMs raised their performance on popular tasks rapidly, their performance on science tasks does not increase.}
    \label{fig:intro_trend}
\end{figure}

Besides presenting a strong model, Intern-S1 represents a step forward in our exploration of finding a viable path toward Artificial General Intelligence (AGI). Figure~\ref{fig:intro_trend} shows that while recent models have made significant improvements in math and general reasoning, they still struggle in science domains that have relatively fewer data. Even if the community can further accelerate the advance of open-source models, their capabilities do not grow evenly across different domains, it's challenging to develop an intelligence system in general domains. 

We believe it's important to discuss the problem, \textit{How can we enhance a model’s capability to tackle low-resource tasks in a scalable way?} Note that the scalability is essential. Unlike in popular domains, we can not heavily rely on heuristics and priors for every low-resource task. Thus, we tackle this problem from more scalable perspectives in pre-training and post-training stages, respectively.

In the pre-training stage, the key challenge is to prepare large-scale pre-training data for those low-resource but high-value science domains. To curate high-quality scientific data, we adopt two pipelines: (1) a recall and filtering pipeline to mine pre-training data from web data with agent workflows, ensuring knowledge coverage. This effort raised the data purity of targeted domains from around 2\% (scientific data rarely occurs in the web-crawled data) to over 50\% according to human evaluation. (2) PDF documents are a rich source of scientific knowledge, and we adopt a page-level PDF document parsing pipeline to obtain high-quality parsed documents at a moderate cost by carefully organizing low and high-cost parsers in the pipeline. These pipelines contributed over 2.5 trillion tokens of scientific data to Intern-S1’s continued pre-training. 

After pre-training, we conduct offline and online reinfocement learning (RL) 
based on InternBootCamp, a large-scale interactive environment designed for foundation models~\citep{li2025internbootcamp} that contains more than 1000 kinds of tasks.
To synergize the simultaneous learning of thousands of tasks with diverse feedback forms in RL, we propose an innovative algorithm framework called Mixture-of-Reward (MoR), which harmonizes the feedback of various forms and tasks into a unified reward scalar.
For hard-to-verify tasks such as creative writing and chatting, the framework adopts POLAR~\citep{dou2025pre} to uniformly provide a reward scalar, implying the distance from the current response to the expected distribution. For various easy-to-verify tasks, it adopts different combinations of verification models~\citep{liu2025compassverifier}, rules, and environmental feedback to generate a reward scalar that precisely indicates the accuracy.
This flexible and targeted design of the reward mechanism endows MoR with higher efficiency, scalability, and adaptability in handling diverse tasks.
We further integrate MoR with multiple techniques of RL algorithms~\citep{cui2025entropy, liu2025cpgd}, and infrastructure optimizations, to stabilize and accelerate large-scale MoE training. 
As a result, we can incentivize the model to learn professional skills using fewer training samples, achieving state-of-the-art performance with 10x less RL training time compared to recent work~\citep{chen2025minimax}. 

Equipped with diverse data curation strategies, high-efficiency infrastructure, and advanced algorithms, \textbf{Intern-S1} achieves state-of-the-art performance among contemporary open-source models and is competitive with—sometimes surpassing—leading closed-source systems (\textit{e.g.}, OpenAI o3, Gemini-2.5-Pro, Grok-4) on our evaluated benchmarks. Intern-S1 excels across a broad suite of scientific-reasoning benchmarks in both text-only and multimodal settings, while maintaining top-tier performance on general-reasoning tasks. We open-source the model weights and accompanying toolchains; this generalist–specialist, integrated design is intended to catalyze future exploration in general reasoning and enable substantive advances across diverse science-focused scenarios.

\section{Model Architecture}
\begin{figure}
    \centering
    \includegraphics[width=0.9\linewidth]{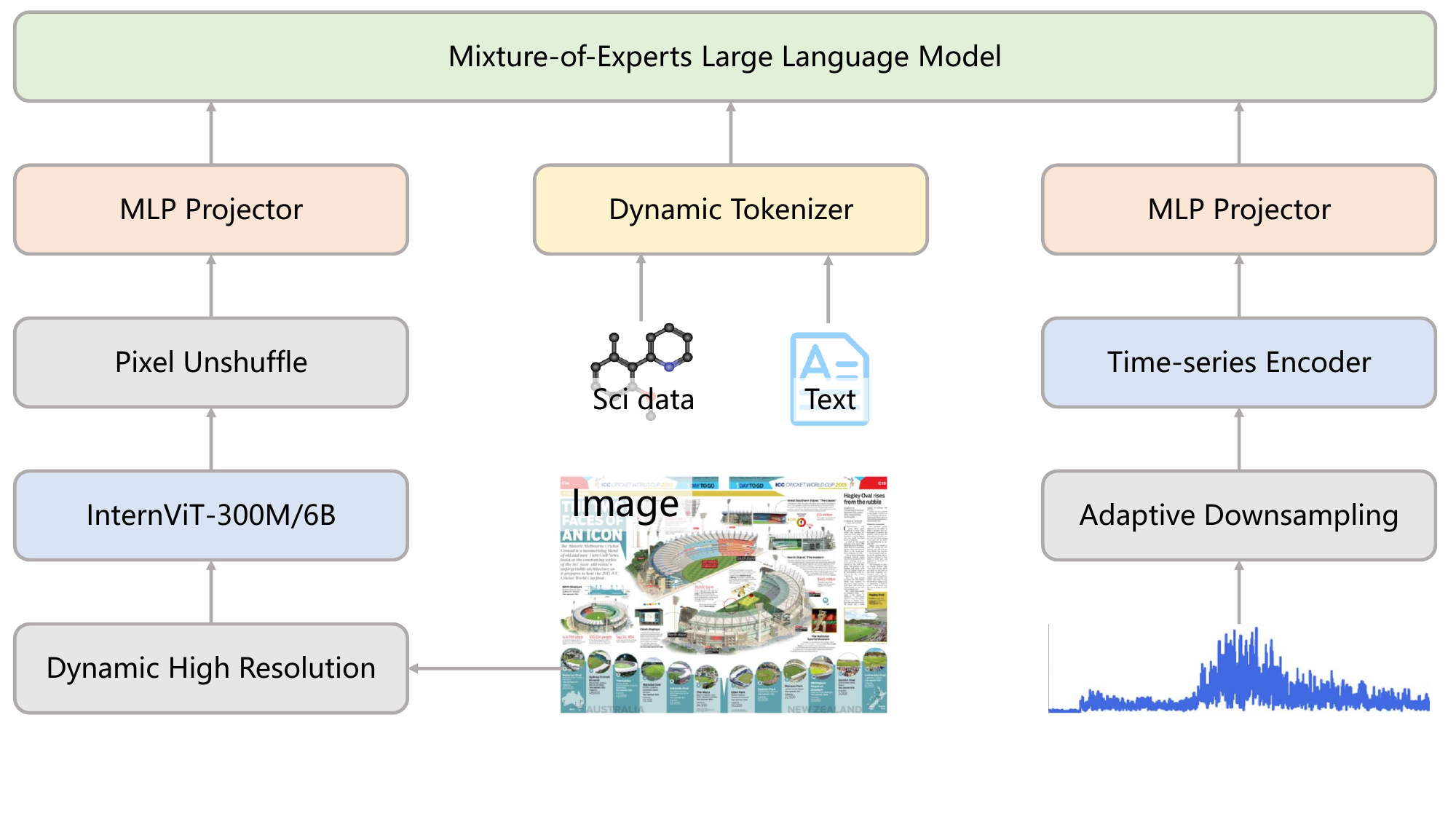}
    \caption{Architecture of Intern-S1, consisting of a MoE LLM with a vision encoder, a time-series encoder, and a dynamic tokenizer that switches the tokenization and embedding strategies for natural language and scientific inputs. The Intern-S1 is equipped with the InternViT-6B, and the Intern-S1-mini is equipped with the InternViT-300M for the consideration of efficiency.}
    \label{fig:main_arch}
\end{figure}

The architecture of Intern-S1 is shown in Figure~\ref{fig:main_arch}.
For the large language model (LLM), we adopt the Qwen3-235B Mixture-of-Expert (MoE) model and Qwen3-8B in Intern-S1 and Intern-S1-mini, respectively.
Based on LLM, we categorize scientific modalities into three types according to their representations and adopt different strategies to project them into the representation space of LLMs. Specifically, we adopt a Vision Transformer (ViT) to encode visualizable representations (\textit{e.g.}, meteorological images), propose a novel dynamic tokenizer for linearizable discrete representations (\textit{e.g.}, molecular structures), and utilize specific designed encoder for domain-specific representations (\textit{e.g.}, time series signals).

\subsection{Vision Encoder}

We employ the InternViT series~\citep{chen2024internvl} as our vision encoders: Intern-S1 uses the large ViT-style InternViT-6B, which is incrementally refined from contrastive pre-training to LLM-coupled next-token prediction, yielding strong high-resolution, fine-grained visual representations.
For Intern-S1-mini, we adopt its compact counterpart, InternViT-300M, a distillation of the 6B teacher further trained with NLP loss, providing an efficient encoder that preserves much of the teacher’s recognition and localization ability as well as its visual world knowledge.
Together, these encoders enable a compute–accuracy trade-off: InternViT-6B maximizes representational power, while InternViT-300M offers a favorable efficiency–performance balance for downstream multimodal tasks.

These encoders can operate at a fixed input size of 448$\times$448 pixels or dynamic resolution~\citep{chen2024far} to better handle high-resolution content. To incorporate with language model, we adopt pixel unshuffle to shrink the number of visual tokens by a factor of four, meaning that a 448×448 image is represented by 256 visual tokens. These visual tokens are then passed through an MLP projector to align them with the language model’s embedding space. The parameter of ViT is joint trained during the image-text training stages to further improve its visual and scientific modalities understanding abilities.

\subsection{Dynamic Tokenizer}

Inspired by prior work~\citep{xia2025naturelm}, we formalize scientific data structures, including molecular formulas and protein sequences, as tagged sequences. For instance, \verb|<SMILES>C1CCCCC1</SMILES>| represents a molecule in the SMILES format. Previous studies have demonstrated that such tagged sequences help language models distinguish between data structure types within one model. However, two important issues remained: (1) the tokenizer applies the same splitting strategy across all sequences, and (2) the same token across different modalities shares the same embedding.

The first issue limits the model’s ability to achieve higher compression ratios in scientific domains. Take SMILES format as an example. Although it is widely-used in chemistry domain, it rarely occurs in general textual corpora. As a result, general-purpose LLMs are inefficient at encoding SMILES format. Since the static tokenizer uses one splitting strategy across all situations, favoring scientific modalities often comes at the cost of reduced compression ratios for natural language text.

The second issue conerns the shared embeddings. For instance, if the character ``C" appears in a DNA sequence, a molecular formula, and a multiple-choice question, forcing it to share the same embedding may bias its representation toward the most frequent usage. This, in turn, can limit the performance in scientific modalities. Although a high dimensional embedding space allows the model store multiple semantic representations in one vector, this frequency imbalance hinders the model from learning those representations precisely. 

Previous studies ~\citep{feher2024retrofitting} have mentioned the limitations of static tokenizers and proposed general-purpose dynamic tokenizers. However, the relevant studies are still in early stages and often suffer from robustness issues, such as the splitting strategy may be sensitive to small contextual changes. Proposed remedies still exhibit limitations like slower convergence speed than standard tokenization methods ~\citep{provilkov2019bpe}. Interestingly, we find that these limitations can be largely mitigated in the scenarios of processing scientific modalities. This is because scientific strings (\textit{e.g.}, SMILES or FASTA format), can be precisely and easily identified, which circumvents the issue of contextual sensitivity.

\begin{figure}[h]
    \centering
    \begin{subfigure}[b]{0.65\textwidth}
        \centering
        \includegraphics[width=\textwidth]{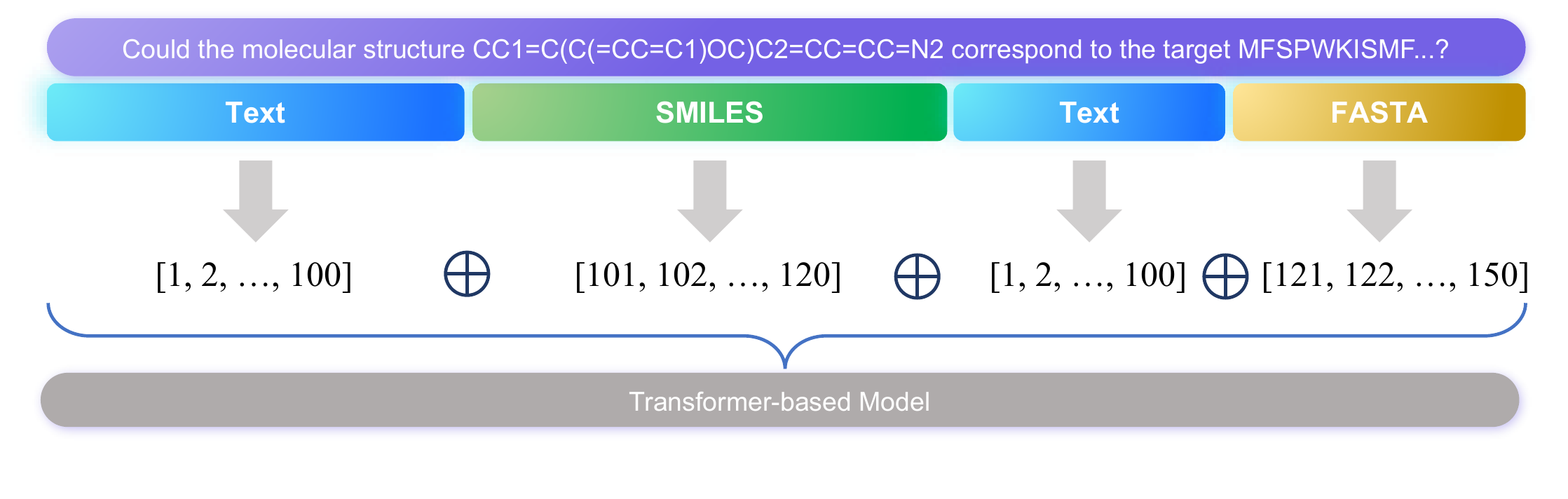}
        \label{fig:sub2}
    \end{subfigure}
    \hfill
    \begin{subfigure}[b]{0.32\textwidth}
        \centering
        \includegraphics[width=\textwidth]{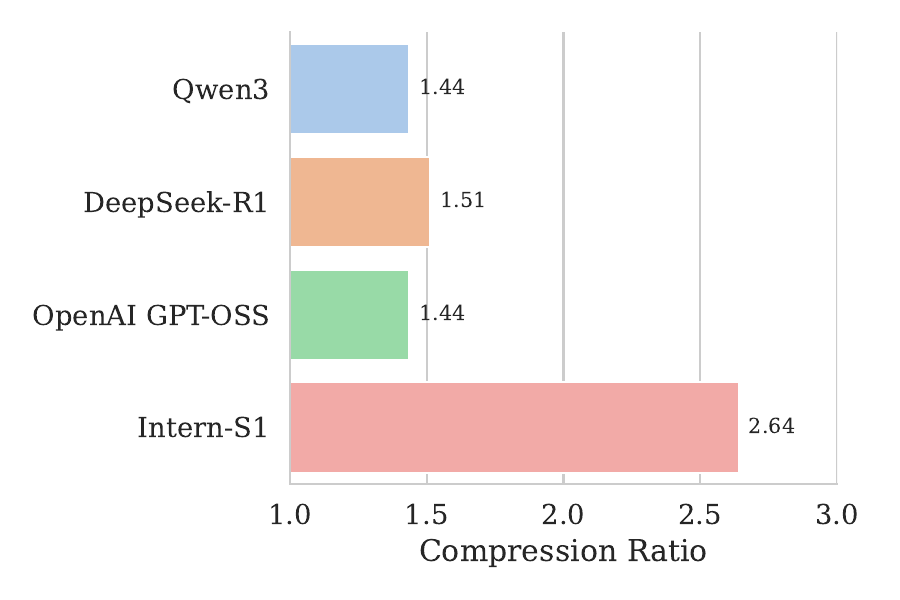}
        \label{fig:sub1}
    \end{subfigure}
    \caption{\textbf{Left}: The workflow of the dynamic tokenizer. The tokenizer will first detect the patterns in the input string using a rule-based detector or user-annotated special tags. Then, it will segment the input string into different parts. Each part will be tokenized using different strategies, and its embedding space will be orthogonal to each other. Finally, those vectors will be concatenated as a regular transformer input. \textbf{Right}: The compression ratio of different tokenizers on scientific data (SMILES format). Intern-S1 outperforms others over 70\%, meaning that the Intern-S1 represents the scientific data with much fewer tokens, saving the computation overhead. }
    \label{fig:tokenizer_combined}
\end{figure} 

The workflow of dynamic tokenizer is illustrated in Figure~\ref{fig:tokenizer_combined}. The tokenizer first identifies the modalities within the input string and then applies different splitting strategies for each. The resulting sequences are concatenated into a single input sequence, maintaining compatibility with modern LLM architectures.

Intern-S1 currently supports four modalities, with plans to expand support in future iterations. Each modality could be clearly marked using special tags in the user input (\textit{e.g.}, \verb|<FASTA>|, \verb|<SMILES>|). We also employ heuristic rules and domain-specific tools (\textit{e.g.}, RDKit\footnote{https://www.rdkit.org}) to automatically detect molecular and protein strings.

Our experiments demonstrate the effectiveness of the dynamic tokenizer. As shown in Figure~\ref{fig:tokenizer_combined}, it achieves a significantly higher compression ratio, improving by up to 70\% over OpenAI’s GPT-OSS-120B, Deepseek-R1, and Qwen3 series.

We compare the compression ratio (CR) of different tokenizers $\tau$ on a chemical dataset $\mathcal{D}$ containing rich SMILES formatted data. The tokenization efficiency was quantified using the Characters-per-Token, formally defined as:
$$
CR(\tau, \mathcal{D}) = \frac{\sum_{s \in \mathcal{ D}} \text{len(}s)}{\sum_{s\in \mathcal{ D}} \text{len}(\tau(s))} 
$$
where string length was measured in Unicode characters.

\subsection{Time Series Encoder}

Intern-S1 integrates a time series encoder to better handle sequential numerical data where each element typically represents a measurement recorded over time, such as seismic waves, gravitational waves, astronomical light curves, and electroencephalography (EEG) recordings. Such data is often long, continuous, and lacks explicit semantic structure, making it less compatible with large language models. The time series encoder captures temporal dependencies and compresses the input into representations that are more suitable for LLM-based understanding and reasoning.

The encoder directly receives and processes raw signals represented as continuous numerical values. The signals can vary widely in sampling rate (from one sample per day to gigahertz-level), duration (ranging from tens to millions of time steps), and physical semantics. The encoder incorporates a dedicated adaptive downsampling module followed by transformer-based blocks, enabling efficient and unified representation of scientific time-series signals. It serves as a  complement to image modality and enhances the model's ability to understand diverse scientific data. 

\section{Infrastructure}

The training infrastructure of Intern-S1 will be released in XTuner~\citep{2023xtuner}, an efficient and flexible toolkit for LLM pretraining and post-training. We organize the infrastructure into two parts: training with the next-token prediction paradigm, including pretraining and supervised fine-tuning, and the reinforcement learning paradigm.

\subsection{Pre-training and SFT Infrastructure}

\noindent \textbf{Parallelism}:
We utilize Fully Sharded Data Parallelism (FSDP) to distribute model parameters across GPUs for continue pretraining and supervised fine-tuning.

\noindent \textbf{FP8 Training}: Following the approach in DeepGEMM~\citep{liu2024deepseek}, we employ FP8 precision for matrix multiplications (GEMMs) with dynamic scaling applied per tile (1$\times$128). During the forward pass, computations involve tile-wise scaling for inputs and block-wise scaling for weights in GEMM operations. In the backward pass, gradients are computed via two types of GEMMs: (1) those with tile-wise gradients and block-wise weights, and (2) those with tile-wise gradients and tile-wise inputs. Additionally, the vision tower is kept in BF16 precision to ensure training stability.

\noindent \textbf{Kernels}: \textit{(1) Grouped GEMM kernel}: To address the challenge of dynamic group sizes in MoE, which result from variable-length in top-k routing, we employ TMA-Adaptive FP8 Grouped GEMM~\citep{suzhongling2025tmaadaptive} to reduce memory and computational overhead in MoE computations. \textit{(2) Liger-kernel}: We fuse linear and cross entropy layers using fused learning cross entropy kernel in Liger-kernel~\citep{hsu2025ligerkernel}. \textit{(3) Flash Attention}: For training, we utilize Flash Attention-3 with variable-length  support to improve attention computation efficiency.

\noindent \textbf{Variable-Length Balanced Strategy}: 
We identify the significant workload imbalance issue in FSDP with variable-length training, particularly at scale. To address this challenge, we propose a \textit{variable-length balanced strategy}(VLBS): (1) randomly packing documents into buckets while recording maximum sequence lengths, (2) applying a sliding window ($S$) to group buckets, and (3) sorting by maximum length within each window. This approach guarantees balanced computational loads across all ranks, yielding an average 2$\times$ speedup in our training framework.

\subsection{RL Infrastructure}

\noindent \textbf{Parallelism}: 
We employ Fully-Sharded Data Parallel (FSDP) with 1-way Expert Parallelism for RL training. This configuration eliminates inter-expert communication and prevents the explosive memory growth that occurs when a dropless MoE is combined with larger EP degrees on long-sequence training.

\noindent \textbf{FP8 Training and Inference}: To maximize rollout throughput, we utilize FP8 precision for both training and inference. This unified approach accelerates data generation by significantly reducing memory bandwidth pressure and increasing computational throughput.

\noindent \textbf{Colocated Design}: Similar to HybridFlow~\citep{sheng2025hybridflow}, our architecture colocates the training and inference engine on the same set of devices. At the start of each RL step, the model is transparently redistributed from its training mesh to the rollout mesh; after collecting the required trajectories, it is redistributed back, with optimizer states intact. Lightweight redistribution and collective synchronization keep memory clean and maintain weight consistency without resource partitioning.

\noindent \textbf{Rollout:} We perform inference with 8-way Expert Parallelism (EP8) using LMDeploy~\citep{2023lmdeploy}. The serving backend is implemented in PyTorch and stores weights in FP8 for minimal memory footprint. CPU off-loading and continuous batching are both enabled to maximize throughput. To prevent stragglers during the length-uncertain decode phase, we re-balance slots on-the-fly whenever the per-rank workload diverges.

\section{Continue Pre-training}

\begin{figure}[h]
    \centering
    \includegraphics[width=1.0\linewidth]{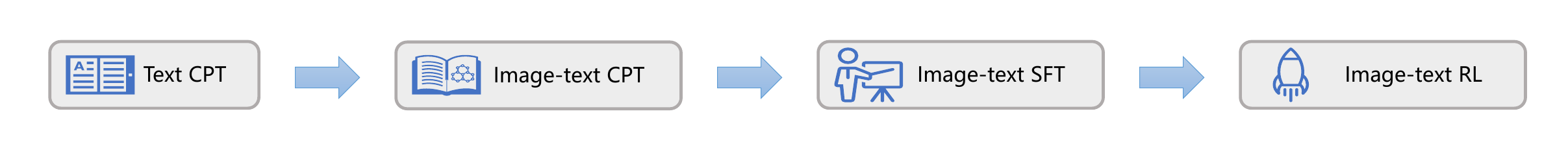}
    \caption{There are four stages for training Intern-S1, and only the first stage is training in the single modality.}
    \label{fig:text_cpt_stages}
\end{figure}

The entire training procedure for Intern-S1 consists of four distinct stages, as depicted in the pipeline in Figure~\ref{fig:text_cpt_stages}. Each stage serves a distinct purpose, characterized by specific types of data preparation and training strategy. In this section, we will elaborate on the details.

\subsection{Scientific Data} 

\begin{figure}[h]
    \centering
    \includegraphics[width=1.0\linewidth]{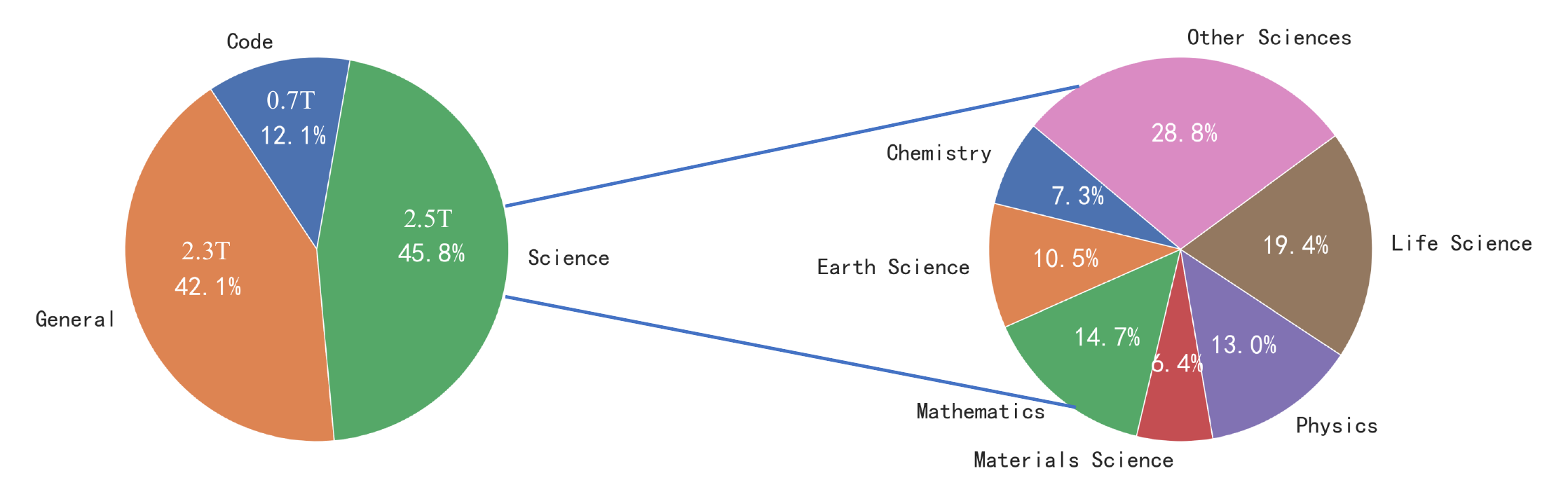}
    \caption{Overall statistics of text CPT data. \textbf{Left}:We continued to pre-train the Intern-S1 on 5T high-quality text tokens in total, and the scientific data occupies over 2.5T tokens. \textbf{Right}: The pie chart illustrates six scientific domains that we spent more attention on and adjusted their distribution in data construction. For example, we adopt strict filtering for life science data and loose filtering for materials science since their natural distributions differ by orders of magnitude.}
    
    \label{fig:text_cpt_data_stat}
\end{figure}

\subsubsection{Text Data}
\label{cha:text data}
During the text and image-text continue pre-training (CPT) process, we sample data from a data pool consisting of high-ratio scientific data. The statistics is shown in Figure~\ref{fig:text_cpt_data_stat}. As the basic text corpus processing pipeline has been well described in other open-source models' technical reports, such as Qwen-series~\citep{yang2025qwen3}, Deepseek-series~\citep{liu2024deepseek}, we highlight three parts of our pre-training data pipeline. 

\paragraph{Page-level PDF documents parsing}
We collect PDF documents from the web and archived libraries, then take a hybrid OCR and VLM pipeline to convert documents into an image-text or pure text corpus. Our lesson is that the parsing quality is crucial to PDF documents, and the PDF documents quality largely impacts the model performance, especially for scientific domains, since they contain extensive knowledge that are rarely mentioned in the web data. We also find that the parsing quality issues mainly related to documents with more equations and symbolic markers. 

Based on human evaluation, we find that none of the existing parsing tools (proprietary or non-proprietary) can perfectly handle all types of PDF documents, and their cost is also diverse in a large range. To balance the quality and cost, we developed a page-level parsing pipeline, where each PDF page will be parsed by a low-cost parser (MinerU~\citep{wang2024mineruopensourcesolutionprecise}). Then we detect the number of equations, symbolic markers, and other heuristic patterns that the low-cost parser often generates bad cases. Based on the detection results, we feed most questionable pages into a high-cost parser (VLMs like InternVL~\citep{zhu2025internvl3} and Qwen-VL~\citep{bai2025qwen2}). VLMs also have different bad case patterns, and the parsing results will be cleaned by rules or a small LLM and merged with other parsed pages. Note that we adopt a global page-level graph deduplication to remove pages like the copyright or other common content. 

\begin{figure}[t]
    \centering
    \includegraphics[width=1.0\linewidth]{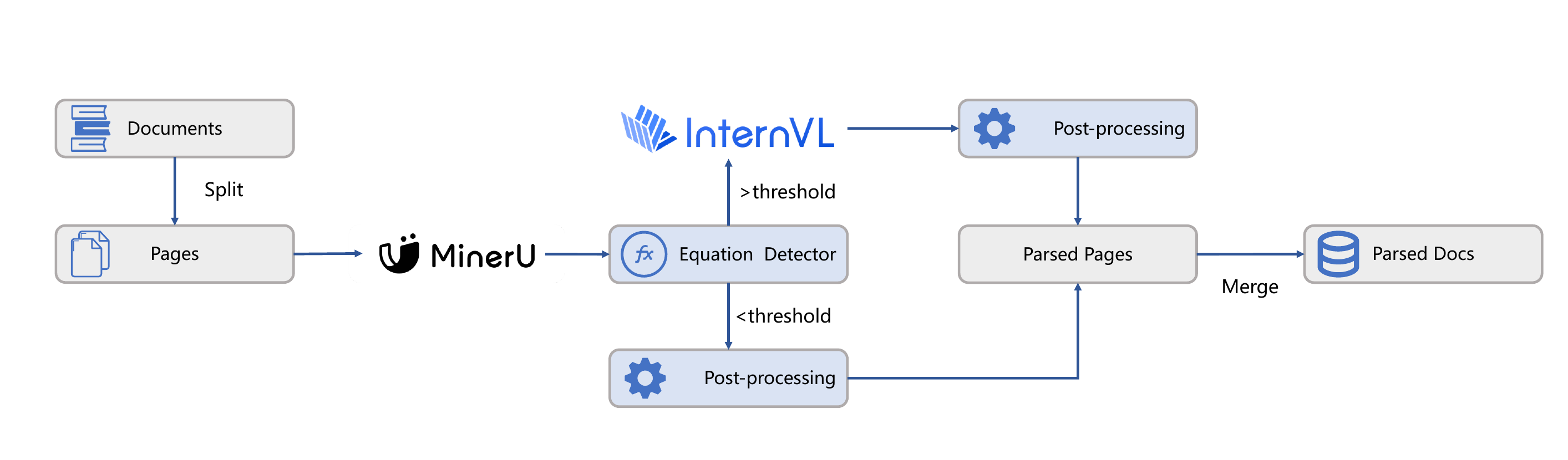}
    \caption{The workflow of our page-level PDF documents parsing pipeline. PDF documents will be split into pages, and a low-cost parser will be used to get textual data. We adopt an equation and symbolic markers detector to check whether a page should go through the high-cost parser (VLMs) for advanced processing or is ready for post-processing. Note that the post-processing for low and high-cost parsers differs since they have specialized bad case patterns. All parsed pages will be merged as a single data sample. }
    \label{fig:placeholder}
\end{figure}

Despite PDF documents being a cleaner source than web data, we still find that quality control is essential. Even with the powerful VLMs, the garble text detection and the page-level deduplication will remove about 20\% of tokens for archived libraries. For web crawling PDFs, we further adopt an education-level scorer that is similar to processing web data, resulting in a 50\% preservation ratio. 

Since the high-cost parser, the VLMs, is 20X slower than the low-cost parser, we only select the pages containing equations and symbolic markers and send them to the high-cost parser. For archived libraries, 5\% of pages have been parsed by the high-cost parser, and for web crawling PDFs, this ratio is 3\%. 

\paragraph{Domain-centric web data parsing}

\begin{figure}[h]
    \centering
    \includegraphics[width=1.0\linewidth]{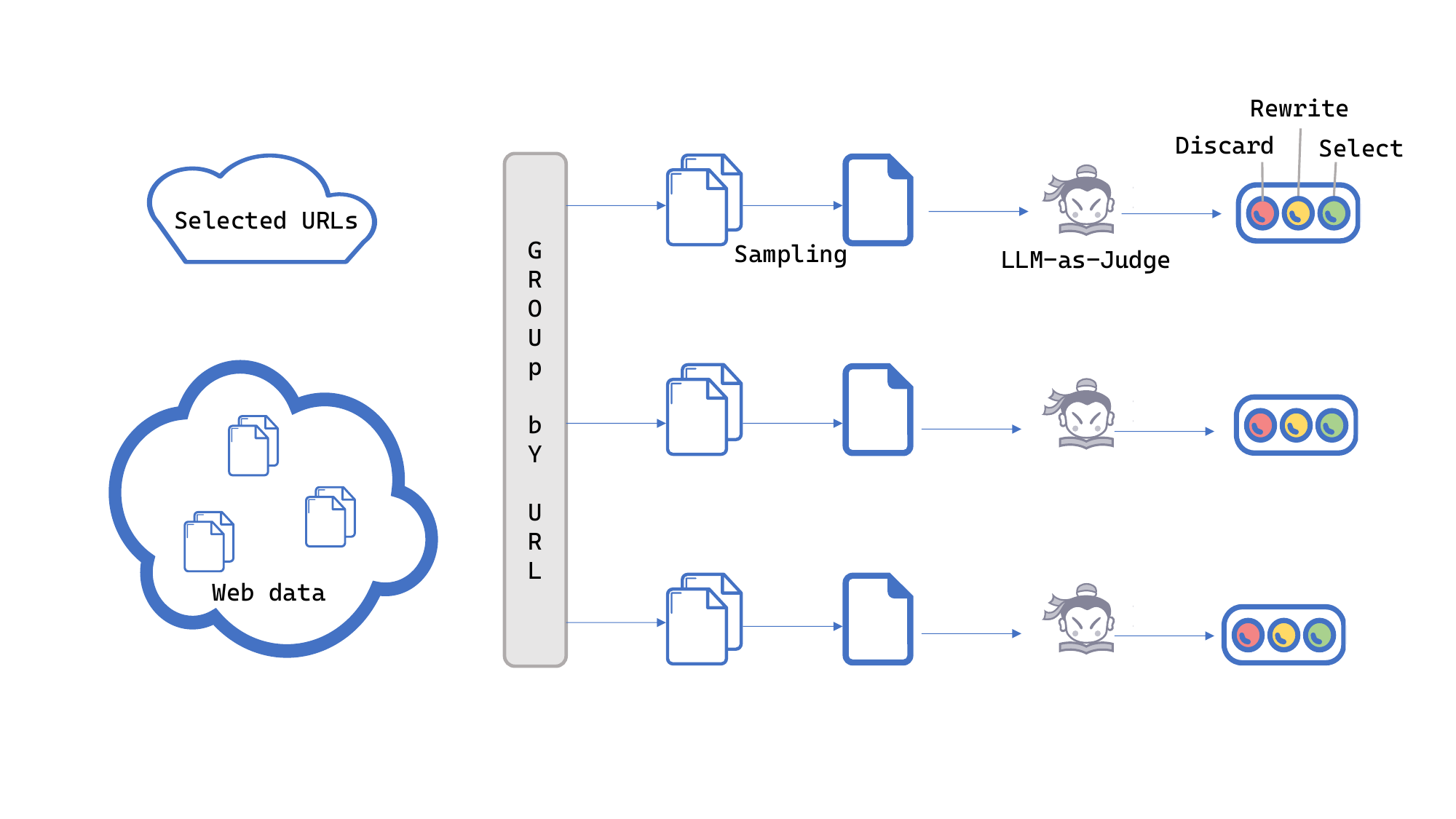}
    \caption{The workflow of our domain-centric web data parsing pipeline. The web pages are grouped by their URL addresses. For each URL domain, we sample hundreds of pages and feed them into an LLM-based classifier. By gathering the classification results of all pages sampled from a URL domain, we make the decision according to heuristic rules. There are three possible actions: discarding all pages from a URL domain if their quality is low and not informative, rewriting all pages from a URL domain using an LLM if their quality is low but the content is informative, and selecting all pages from a URL domain as the training data candidate. }
    \label{fig:placeholder}
\end{figure}

Web-crawled data has been extensively studied as a main source for pre-training data. However, URL domains exhibit diverse and distinct characteristics, making it challenging to design a universal parsing solution for all web pages. Previous works have categorized web pages using topic and format classifiers, and we further introduce a more fine-grained approach, domain-centric web data parsing pipeline. This pipeline treats web pages from the same URL domain as a coherent unit and applies customized strategies for each domain by a LLM-based agent.

Specifically, we sample hundreds of web pages from a given domain and feed them into a high-cost LLM-based agent. According to the tags annotated by this agent, we aggregate the results at the domain level and decide whether to discard, rewrite, or retain pages under that domain. Besides, we also incorporate commonly-used rule-based filtering, deduplication, quality and format classifiers at page-level.

Our motivation of the domain-centric pipeline is that pages from the same URL domain often share common characteristics—such as recurring parsing issues, for example, failed code snippet extraction, or customized navigation bars that are difficult for standard filters to detect. On the other hand, since the LLM-based agent is too costly to apply to all pages, domain-level parsing allows us to recognize structural patterns that lightweight classifiers cannot, while maintaining an acceptable cost.

\paragraph{Scientific data recall and filtering}

\begin{figure}[h]
    \centering
    \includegraphics[width=1.0\linewidth]{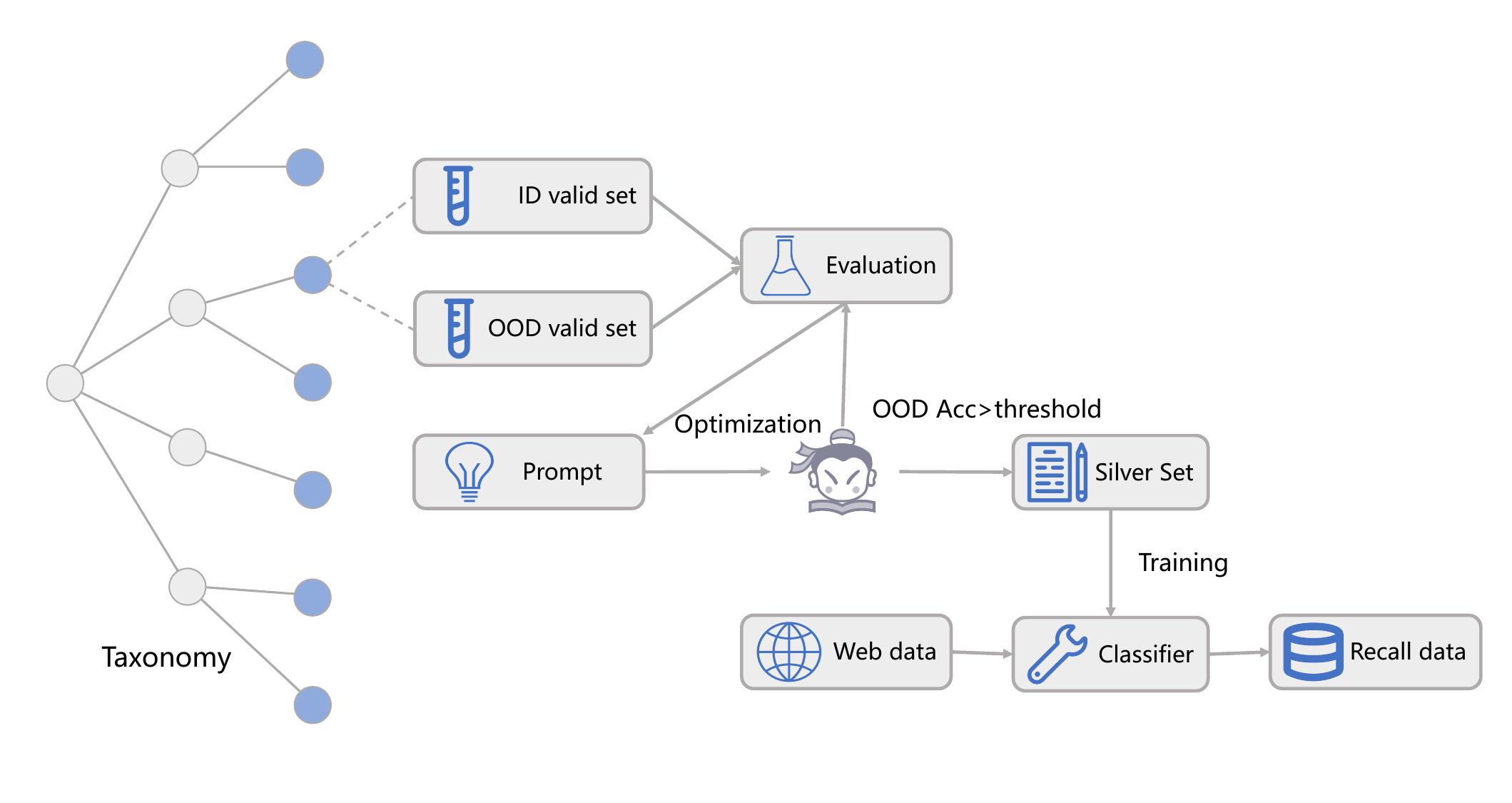}
    \caption{The workflow of our scientific data recall and filtering pipeline. According to a taxonomy that covers various scientific and general domains, we construct a specialized recall and filtering pipeline for each target domain. We prepare the in-domain and out-of-domain validation sets to assist in evolving the prompt automatically. This prompt will trigger the LLM to annotate a large silver set to train the low-cost classifier to filter the web data pool and recall the demand data.}
    \label{fig:text_cpt_recall}
\end{figure}  

To enhance the model's capability in scientific domains, we recall and filter relevant data from open-source pre-training corpus ~\citep{su2024nemotron, chang2024redstone, tang2024txt360} and web-crawled data\footnote{https://commoncrawl.org}. We construct a three-level taxonomy tree of data domains, similar to ~\citep{du2025supergpqa}, and select six scientific domains (Mathematics, Physics, Chemistry, Life Science, Earth Science, and Materials Science) for fine-grained processing.

As shown in Figure~\ref{fig:text_cpt_recall}, for each target domain, we leverage a strong LLM to annotate a subset of data, which serves as the training set for lightweight classifiers (fastText models ~\citep{joulin2016fasttext} and 1.5B-parameter LLMs). We also construct in-domain and out-of-domain (OOD) validation sets for each target domain. The in-domain validation set is drawn from the same source, while the OOD validation set is sourced from a different domain, for instance, using a PDF document as the OOD validation set for web data parsing. These validation sets are used to refine the prompts for training data annotation. As a result, manual evaluation across the six domains shows that the proportion of target domain data increased from 2\% to 50\%, demonstrating the effectiveness of our recall and filtering strategy.

\subsubsection{Multi-modal Data} 
\label{sec: mm cpt data}
In the image–text CPT stage, we curate two categories of datasets: an interleaved image–text dataset and a purely textual dataset. To ensure high data quality, domain diversity, and comprehensive disciplinary coverage, our data sources are drawn from three main origins: 1) Multi-modal pre-training corpus from InternVL3, which spans a wide range of domains including image captioning, general question answering, mathematics, charts, optical character recognition (OCR), knowledge grounding, document understanding, multi-turn dialogue, and medical data.
2) Textual corpus sampled from the dataset described in Section~\ref{cha:text data}, aimed at preserving the model’s text understanding and reasoning capability.
3) Multimodal scientific data, covering specialized domains to enhance performance in expert tasks.

Under this configuration, the total number of training tokens is approximately 250 billion, comprising 70 billion from language data and 180 billion from interleaved image–text data (30 billion tokens for scientific data). 

\paragraph{Multimodal scientific data pipeline.}

Building on the text-only scientific pipeline introduced in~\ref{cha:text data}, we construct a multimodal (image–text) pipeline to 1) preserve fine-grained scientific structure (figures, equations, symbols, tables, charts), 2) align visual assets with surrounding textual context at page/figure/snippet granularity, and 3) produce instruction-style and exam-style supervision suitable for scientific reasoning.

For exam-style problems across six scientific domains, we adopt a filter to check structural integrity. Each sample must contain the question stem, options (if applicable), answers, and explanations; instances with missing fields are discarded. Practically, we adopt rule-based filters to remove unclear stems, incomplete option sets, or answers inconsistent with stems/options. For fill-in-the-blank questions containing multiple sub-questions, we additionally employ large language models such as Qwen2.5~\citep{team2024qwen2} to assess answer completeness.

For the parsed PDF documents and other data containing the latex and markdown equations, we validate the rendering results of symbolic markers to prevent formula corruption and typographic errors based on VLMs' judgments. For general image-text pair data, the basic rule-based filters check the bad cases, including blank images, visibly blurred, distorted figures, or broken links between stems and visual assets.

\subsection{Training Strategy}

\paragraph{Overview of Training Stages}

Intern-S1 adopted a multi-stage training to optimize both effectiveness and training efficiency. As shown in Figure~\ref{fig:text_cpt_stages}, the text-only training stage only involves the text continue pre-training (text CPT), and all other stages are multi-modal joint training, reflecting our strong emphasis on the integration and alignment across different modalities. To achieve the high training efficiency, we adopt a batch size warmup strategy, which can largely keep a good training loss while allowing a large batch size for infrastructure optimization.

\subsubsection{Batch size Warmup}

\begin{figure}[h]
    \centering
    \includegraphics[width=0.7\linewidth]{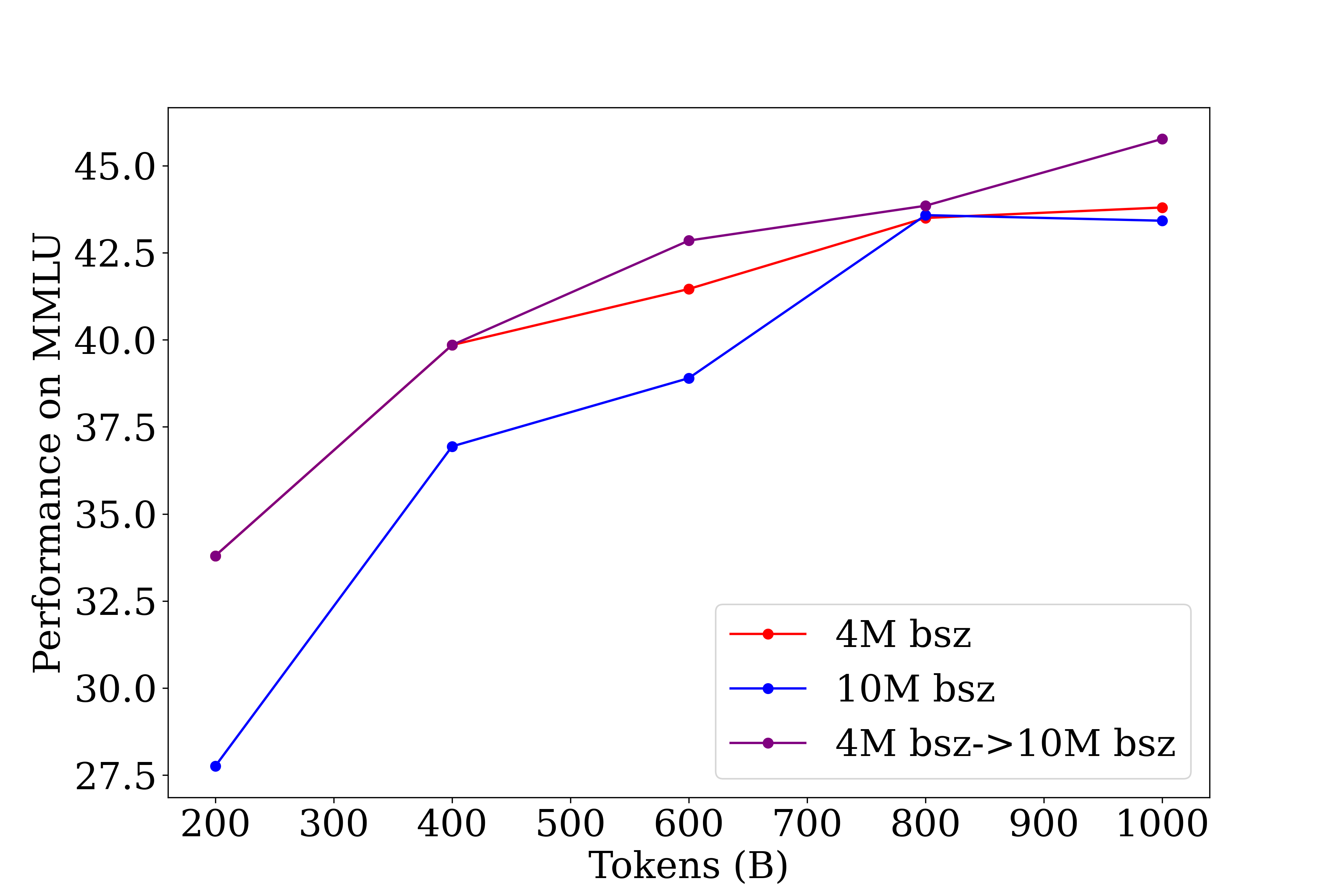}
    \caption{Performance trends of training with different batch size strategies. We train a small LLM with 1B parameters over 1T tokens and select the performance on MMLU as a proxy of the model's downstream performance. The red and blue line used a consistent batch size during the training, and the purple line represents the training process that we switch the batch size from 4M to 10M after training 400B token.}
    \label{fig:text_cpt_bsz}
\end{figure}

Batch size is a crucial hyperparameter that largely influences both the optimization and the infrastructure efficiency. However, this is a dilemma between the optimal model performance and the optimal training efficiency, and it becomes a severe issue at the beginning of the training. As illustrated in Figure~\ref{fig:text_cpt_bsz}, we can see the model trained with a smaller batch size outperforms the model trained with a larger batch size during the early stage (the first 700B tokens). In contrast, using a larger batch size provides a higher training efficiency. Based on our scaling law analysis, we find that the best option is to split the training stage and adopt a different batch size. The simplest version initially takes a small batch size and switches to a large batch for high training efficiency.

\subsubsection{Starting point choice}

To make better use of computational resources and to focus on scientific research, we continue training based on existing open-source models (Qwen3~\citep{yang2025qwen3} and InternVL-Vit~\citep{chen2024internvl}). These models typically have two versions: base model and instruction model, raising the question of which to choose. In short, our study indicates that the instruction model performs slightly better than the base model. While instruction models generally have better performance on downstream tasks, they also tend to produce narrower output distributions. We examine the choice of starting point from two perspectives: 1) Does better initial performance lead to higher final performance? 2) Does initial output diversity affect the RL process?

\begin{figure}[h]
    \centering
    \includegraphics[width=0.8\linewidth]{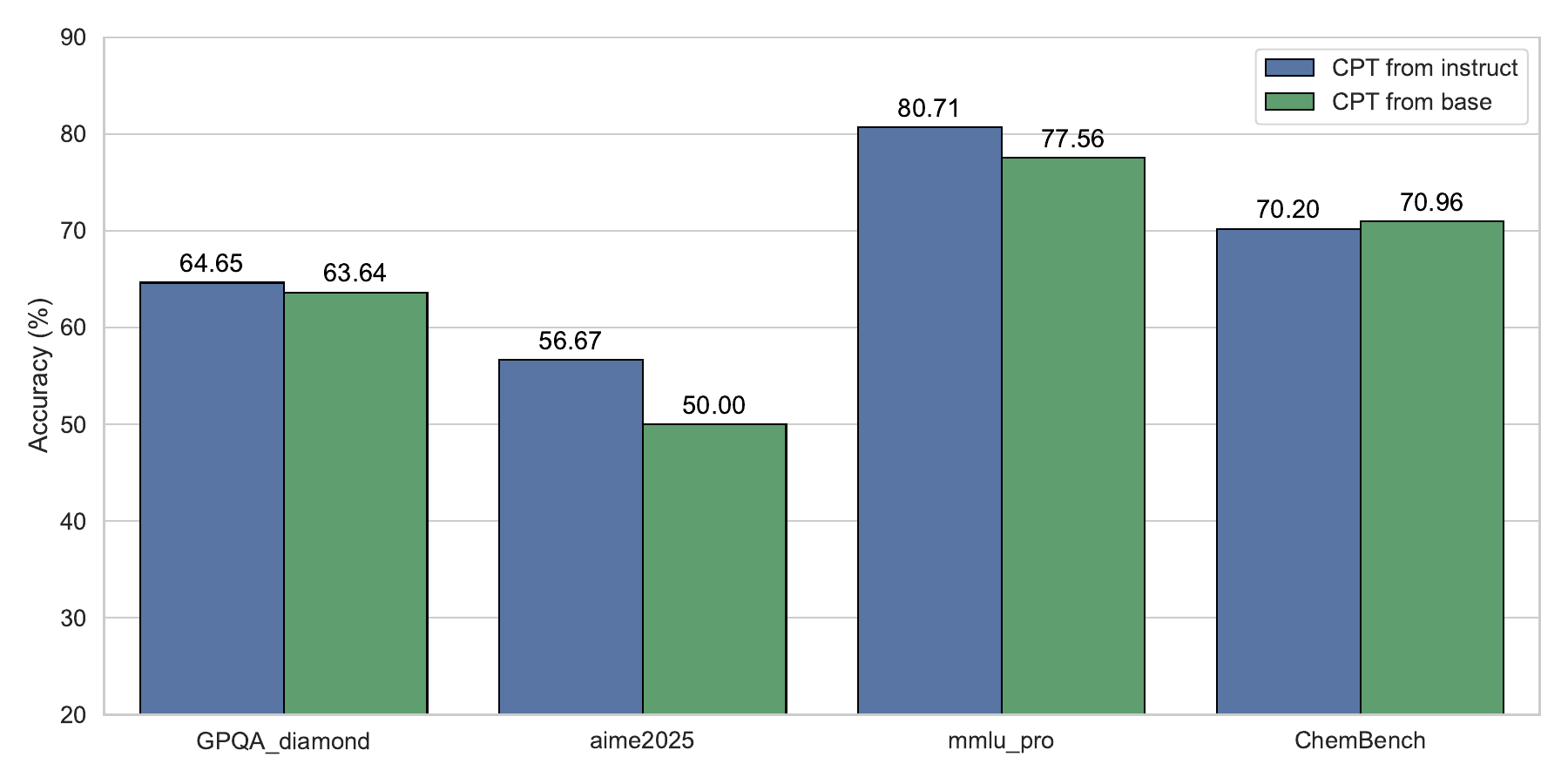}
    \caption{Comparison of choosing different staring points (base and instruction models).}
    \label{fig:text_cpt_starting_point}
\end{figure}

We first conduct experiments using a small model under four settings: 1) directly fine-tuning an instruction model, 2) continuing pre-training and then fine-tuning an instruction model, 3) directly fine-tuning a base model, and 4) continuing pre-training and then fine-tuning a base model. As shown in Figure~\ref{fig:text_cpt_starting_point}, the instruction model only has an advantage on the coding benchmark. According to recent studies ~\citep{ward2025reasoning, dong2025rl}, we think there are two scenarios: 1) post-training activates capabilities already present in the base model. In this case, starting from either the base or instruct model yields similar results. 2) post-training brings new skills and enhances model capabilities. In this case, starting from the instruct model leads to better performance. We guess that the tested instruct model's post-training improves its coding capability, which explains the observed advantage. In contrast, post-training in other domains may only to activate pre-existing capabilities learned during the pre-training, resulting in similar behavior between the base and instruction models.

Another concern is the output diversity when using a base or instruct model as the starting point. More specifically, the initial entropy is crucial to the RL process, as it significantly impacts both the choice of hyperparameters and the potential performance upperbound. We evaluate two settings: 1) directly fine-tuning a base model, and 2) continuing pre-training and then fine-tuning an instruction model. On a subset of our math reasoning RL prompts, the base model shows slightly higher initial entropy compared to the CPT on the instruction model (0.19 vs. 0.15). Based on our experience, this level of difference can often be mitigated through appropriate tuning of RL hyperparameters and is unlikely to result in fundamentally different performance outcomes.

In summary, our findings suggest that:
\begin{itemize}[leftmargin=20pt]
    \item Using the instruction model as the CPT starting point is acceptable in terms of final performance after SFT and RL.
    \item The instruction model is a better choice than the base model when post-training significantly improves model capabilities. However, we observe this effect only in specific domains.
\end{itemize}

\subsubsection{Hyper Parameters}

Within the self-developed Xtuner~\citep{2023xtuner} training framework, we have conducted a fine-grained investigation into the settings of the batch scheduler and learning rate. Initially, the Warmup-Stable-Decay (WSD) paradigm for the learning rate scheduler has gained considerable favor due to its ability to decouple the schedule from the total volume of training data. Following an in-depth study, we have demonstrated that under a gradient-based optimizer, when a WSD learning rate scheduler is employed, for the model to achieve a specific training quality (i.e., for the loss to converge to a certain value), the gradient noise and the batch size must satisfy the following relationship:
\begin{equation}
    1-\frac{1}{2} \eta \frac{\mathcal{B}_{\text{simple}}}{B}>0, \quad \mathcal{B}_{\text{simple}}=\frac{tr(\Sigma)}{|G|^2},
    \label{eq:1}
\end{equation}
where $\eta$ is learning rate, $B$ is the batch size, $\mathcal{B}_{\text{simple}}$ represents gradient  noise, and its meaning can refer to~\citet{mccandlish2018empirical}.

A power-law relationship exists between $\mathcal{B}_{\text{simple}}$ and Loss, as established in~\citet{mccandlish2018empirical}. This implies that throughout the entire pre-training, the batch size coordinated with the WSD learning rate scheduler should follow a gradually increasing progression.
Within the Xtuner framework, after a comprehensive evaluation of both training efficiency and effectiveness, we transitioned the batch size from 66M to 132M tokens.
Furthermore, by utilizing the relationship between the Critical Batch Size and training loss, we not only precisely identified the optimal stage for the transition—switching the batch size after processing 400B tokens under our training configuration—but also ensured maximal computational efficiency throughout the training process.

After addressing the batch size scheduler, the next consideration is the specific learning rate setting under the WSD learning rate scheduler. 
To avoid significant resource waste, it is common practice to identify an ``optimal" learning rate using small-scale resources and then transfer it to the training of large-scale models~\citep{yang2021tuning,bi2024deepseek}. However, such approaches are sensitive to various training settings, such as the learning rate scheduler, batch size, and training data.
To mitigate the influence of these other training settings, we have explored a method for setting the learning rate based on Scaling Laws. Specifically, we first establish the relationship between the training loss $L$ and the learning rate at each step $\Omega=\{\eta_{i}\}$, during the training process, with the concrete expression detailed in~\citet{luo2025multi,tissue2024scaling}. Inspired by the work~\citep{luo2025multi}, after fitting $L(\Omega)$, we formalize the learning rate setting as an optimization problem with respect to $\Omega$. Building on this foundation, we propose a more general optimization expression, as follows:
\begin{equation}
    \min \limits_{\Omega}L_{\theta}(\Omega),\quad s.t. \Omega=\{\eta_{i} | 0 \leq i \leq T, 0 \leq \eta_{i} \leq \mu  \},\quad \phi(\Omega),
    \label{eq:2}
\end{equation}

where $\phi(\Omega)$ represents the constraints on the learning rate. By solving this optimization problem, we ultimately determined the learning rate for the entire pre-training process.

\paragraph{Predictable Loss}
Throughout the entire text CPT process, based on the efficient adaptation with the self-developed Xtuner training framework, rigorous control over training data quality, and the scientific configuration of hyperparameters, the quality of the Intern-S1 pre-training was well-controlled. The entire process did not encounter any loss spikes. Furthermore, based on the pre-training data used for Intern-S1, the final training loss was predicted to be around 1.16 according to the fitted Scaling Laws. In practice, the final training loss for Intern-S1 landed between 1.17 and 1.18, demonstrating our ability to control pre-training quality with a precision of up to the 0.02 level.

\subsubsection{Multi-modal Training}

After the text CPT stage, we conducts integrated optimization over interleaving multimodal data (\textit{e.g.}, image–text pairs, video–text pairs, or interleaved image–text sequences) and large-scale textual corpora.

Conventional approaches often freeze certain layers of the LLM component, or even the ViT encoder, when adapting to MLLMs. In contrast, our method updates all model parameters jointly during multimodal CPT process. Let $\mathcal{M}$ denote a Transformer-based model parameterized by $\theta$ that is capable of processing text, images, and videos in a unified manner. Given an arbitrary training sample $\mathbf{x} \;=\; \bigl(x_1, x_2, \dots, x_L \bigr)$  with token length $L$, we employ the standard left-to-right autoregressive objective:

\begin{equation}
    \mathcal{L}(\theta)
    \;=\; - \sum_{\substack{i=2 \\ x_i \,\in\, \mathrm{Text}}}^{L}
    w_{i} \cdot
    \log \, p_\theta \bigl(x_i \,\bigm|\; x_1, \dots, x_{i-1}\bigr).
\label{eq:autoregressive-loss-text}
\end{equation}

where $w_i$ denotes the loss weight for token $i$. 
Following InternVL3~\citep{zhu2025internvl3}, we adopt a square-averaging scheme to mitigate gradient bias. Specifically, for a training sample with $l$ tokens contributing to the loss, we set 
\begin{equation}
  w_i = l^{-1/2}
\end{equation}

Visual tokens are used solely as conditioning context for text prediction and are not themselves predicted. This formulation encourages the model to integrate multimodal information in a way that benefits downstream language generation tasks.


\section{Post-training}

After multi-modal continued pre-training (CPT), Intern-S1 undergoes a two-stage multi-modal post-training, where text data and multi-modal data are mixed to obtain optimal performance in both text and multi-modal benchmarks.

\subsection{Offline Reinforcement Learning}

In the post-training stage of Intern-S1, we first conduct offline reinforcement learning (RL) on meticulously curated instruction data, where each response to the training query is obtained from best-of-N (BoN) sampling based on the corresponding criteria such as accuracy, fluency, and safety. Based on the curated instruction data of various domains, we further conduct data mixture experiments to obtain the best data mixtures at this stage.
Although this stage is more conventionally called supervised fine-tuning (SFT), we treat this stage as offline RL to highlight that all the used responses are essentially rewarded due to BoN sampling.

\subsubsection{Instruction Data Curation}\label{sec:sft-general-data}

\paragraph{Text-only Data.}
Extensive research~\citep{li2023quantity,liu2023makes,shen2024rethinking} has demonstrated that the quality and diversity of instruction tuning data are of paramount importance. A small number of low-quality or harmful data samples can lead to model collapse or the acquisition of undesirable patterns, while more diverse and balanced data often yield superior model performance. Therefore, we develop a data pipeline to ensure both diversity and high quality of the instruction data. For Intern-S1, the instruction distribution is meticulously designed, and extensive ablation experiments are conducted to derive the optimal training dataset.

The data pipeline primarily consists of three components: \textit{Filtering}, \textit{Labeling}, and \textit{Enhancement}.
For filtering part, we employ a combination of rule-based and model-based filtering to eliminate any data that could potentially hinder model training, such as repetitive expressions, truncated data, or hallucinated content. 
Based on the filtered data, we further implement a labeling strategy to facilitate subsequent data selection. Labeling is performed along two key dimensions: \textit{data category} and \textit{data difficulty}.
\begin{itemize}
    \item For data categories, we manually predefine a hierarchical labeling system comprising at least three levels (\textit{e.g.}, \textit{Mathematics → Advanced Mathematics → Linear Algebra}). A labeling model is then used to classify all data into their respective category labels. Additionally, the model extracts specific contextual information to generate scenario labels (\textit{e.g.}, \textit{Linear Algebra Proof Exercise}). 
    \item 
For data with ground truth, we use a small-scale model to perform multiple rollouts, using the pass rate to determine difficulty. For data without ground truth, domain-specific difficulty criteria are established, and the labeling model assigns difficulty levels accordingly. 
\end{itemize} 
Based on the category and difficulty of each data sample, we perform stratified sampling to maintain a balanced distribution across domains and difficulty levels.  
Finally, in cases where certain domains suffered from insufficient data volume or an overabundance of low-quality samples, we either reconstruct the responses or employ synthetic data generation~\citep{cao2025condor} to supplement the dataset.

Through this pipeline, we obtain a candidate data set that spans multiple categories, including \textit{Agent, Code, General Dialogue, Instruction Following, Mathematics, Reasoning, Long Text, Safety, Chemistry, Life Sciences, and Physics}. We conduct extensive ablation experiments using this candidate dataset to obtain the best mixtures.

\paragraph{Multi-modal Data.}
The initial vision-language (VL) instruction data used for Intern-S1 is the same as that used by InternVL3~\citep{chen2024internvl}, which includes specialized capability data for 3D scene understanding, GUI manipulation, long-context reasoning, video comprehension, scientific diagrams, and creative writing.
To facilitate the long-thinking capability of Intern-S1 given visual inputs, we further construct vision-language reasoning data by enhancing the pipeline of SOPHIA~\citep{shen2025semioffpolicyreinforcementlearningvisionlanguage} with more strict quality controls (\textit{e.g.}, rejected sampling, de-duplication, length/format constraints, self-consistency checks, and programmatic verification when available). 
Based on that, we further include the VL instruction data in scientific domains to bolster the competence of Intern-S1 in scientific reasoning.

\subsubsection{Data Mixture Experiments}

Based on the curated text and multi-modal instruction data, we still need to determine the optimal mixture ratio to obtain balanced performance across general language and multimodal understanding and reasoning, and scientific reasoning benchmarks. The experiment is conducted in two main steps: 

\paragraph{Atomic Capability Validation} First, we validate the effectiveness of each incremental dataset. To accelerate ablation experiments, we first perform proportional sampling from the SFT data of InternLM3 to curate a core dataset serving as the baseline for ablation studies. Building upon this foundation, we incrementally introduce additional domain-specific data subsets while monitoring corresponding improvements in benchmark performance. For instance, we jointly train the model on a newly added math dataset alongside the core dataset and evaluate whether it leads to performance gains on the relevant math benchmark.

\paragraph{Compositional Capability Validation} After validating the efficacy of all individual data components, we proceed with comprehensive evaluation by merging these datasets. This integration phase addresses inter-domain data conflicts and involves hyperparameter tuning to ultimately determine the optimal training configuration, including both hyperparameters and data mixture ratios. Specifically, we begin by sampling and training on domain-specific data using initial heuristic ratios. We then refine the proportions based on benchmark performance and mitigate data conflicts via techniques such as style alignment and curriculum learning, ultimately deriving the optimal data mix and hyperparameters. 

During training, we also adopt the random JPEG-compression augmentation and the squared-loss objective introduced in InternVL3~\citep{zhu2025internvl3}. We set the maximum context length to 32K tokens, thereby mitigating truncation effects, enhancing the model’s ability to capture long-range dependencies, and improving performance on document-level and multi-image reasoning tasks. 

\subsection{Online Reinforcement Learning}

\subsubsection{Mixture-of-Rewards Framework}

\begin{figure}[htbp]
    \centering
    \includegraphics[width=0.7\linewidth]{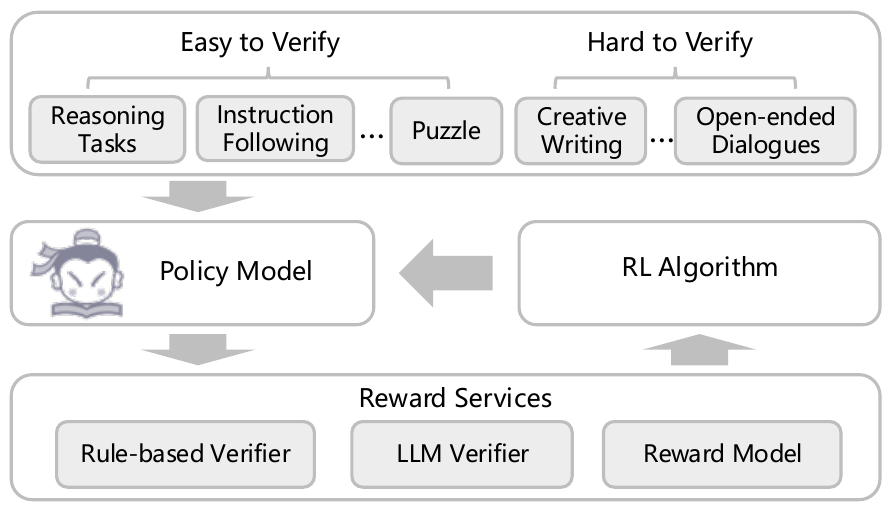}
    \caption{The Mixture-of-Rewards framework. }
    \label{fig:MoR}
\end{figure}

The second stage of post-training is online reinforcement learning, which covers the simultaneous learning of more than 1000 tasks. To achieve that, Intern-S1 employs a Mixture-of-Rewards (MoR) framework (Figure~\ref{fig:MoR}) that integrates diverse reward signals from multiple multimodal tasks, enabling the model to acquire both domain-specialized and general-purpose capabilities within a unified optimization process.
Rather than optimizing for individual capabilities in isolation, MoR harmonizes diverse reward signals from heterogeneous task types such as logical reasoning, domain-specific academic problems, instruction-following tasks, visual-textual reasoning questions, and general conversation.
This cross-task policy optimization enables the model to maintain high performance on specialized tasks while preserving robustness and adaptability in open-ended dialogue scenarios.

However, during mixed training across multiple tasks, challenges arise, such as varying difficulty levels and divergent convergence speeds among different tasks. 
To address these issues, we implement a hybrid offline–online data filtering strategy for RL training. 
Initially, we conduct separate rollouts and training for each task domain to assess their respective difficulty levels and convergence speed. Based on these measurements, we determine the final data mixing ratio. 
During the training process, we roll out trajectories with a number of prompts greater than the training batch size and perform online filtering based on criteria such as data accuracy and output quality. 
The generation process stops once the amount of generated data meets the training batch size. 
This combined offline and online strategy not only balances the performance across multiple tasks, preventing overfitting on some tasks while underfitting on others, but also enhances the overall training efficiency.

\subsubsection{Tasks and Verifiers}\label{sec: tasks}
\paragraph{Large scale sandboxes} 

In Intern-S1, we utilize InternBootcamp~\citep{li2025internbootcamp} to provide a wide range of large-scale synthetic reasoning tasks, including algorithms, character reasoning, cryptography, graphical puzzles, boardgame reasoning, logical reasoning, and science scenarios such as physics equations, chemistry formulas, and medical reasoning.
In total, we have over 1,000 different tasks with unlimited training data using the bootcamp case generator.
These tasks apply the InternBootcamp framework and provide verifiable rewards for the mixture of rewards.
For each task, we craft over 100,000 training samples through the data generator, and we downsample these different samples by heuristic rules.
In total, we have over 20,000 samples covering different scenarios used in the mixed RL training process.

\paragraph{Instruction following} 

Instruction-following capability, serving as the foundation for model generalization, requires verifiable reward signals for targeted optimization. We adopt the dataset from \cite{guo2025ifdecorator}, which offers instruction data with a balanced difficulty distribution. To exclude trivial and overly difficult cases, we compute passrate@64 scores using the SFT model and select prompts with scores in the range [0.2, 0.8]. Driven by this data-reward mechanism, the model exhibits sustained enhancement in instruction-following \cite{zhou2023instruction} performance.

\paragraph{Multi-modality reasoning tasks} 
To enhance the reasoning capabilities of Intern-S1, we construct a high-quality multi-task, multi-modal dataset for Reinforcement Learning with Verifiable Rewards (RLVR). 
For textual data, we focus on strengthening mathematical reasoning, using questions from open source datasets, such as OREAL-RL-Prompts~\citep{lyu2025exploring}, DAPO-Math-17k~\citep{yu2025dapo}, and Skywork-OR1-RL-Data~\citep{he2025skywork}, as well as internal data at the university level and the competition level. 
For multimodal data, we collect samples from diverse domains, including general visual question answering (VQA), science reasoning, chart question answering, mathematical reasoning, document understanding, and OCR, sourcing from MMPR~\citep{wang2024enhancing}, MMK12~\citep{meng2025mm}, and private collections. 
Additionally, we apply rendering techniques to convert part of the pure-text questions into image format for training. 
To eliminate the noise of random guessing during RL, we reformat some multiple-choice questions into a fill-in-the-blank format. 

In multi-modality reasoning tasks, all questions have verifiable ground truth answers.
We employ the CompassVerifier~\citep{liu2025compassverifier} as a generative lightweight verifier for evaluation and outcome reward for multi-domain competency spanning math, knowledge, and diverse reasoning tasks, combined with a rule-based verifier, to evaluate the correctness of the outputs and provide binary rewards. This combination enhances the robustness of correctness assessment, mitigating issues like the false negatives of the rule-based verifier.

\paragraph{Human preference alignment} 
To align Intern-S1 with human preferences in open-ended scenarios, we employed POLAR-7B ~\citep{dou2025pre} as the primary reward function. POLAR-7B is trained under the paradigm of Policy Discriminative Learning, which enables the model to discern identical policies and discriminate between different ones. In contrast to traditional reward modeling methods that rely on absolute preferences, POLAR captures the relative difference between two policies. This approach provides a scalable, high-level optimization objective well-suited for modeling generic ranking relationships. The pre-training of POLAR-7B utilized a large-scale synthetic corpus of 3.6T tokens, followed by fine-tuning on 150K preference pairs with references.

For the RL stage, we construct a dataset whose prompts are sourced primarily from established open-source instruction datasets, including UltraFeedback~\citep{cui2023ultrafeedback} and HH-RLHF~\citep{bai2022training,ganguli2022red}, and augmented with a subset of anonymized, real-world user queries. A reference trajectory for each prompt was generated by a randomly selected SOTA LLM from a pool that included both open-source and close-source models. This methodology enabled the creation of a high-quality, large-scale RL dataset entirely through automated means.

\subsubsection{Policy Optimization}

Recent research in long chain-of-thought reinforcement learning has demonstrated notable gains in reasoning tasks, especially GRPO~\citep{shao2024deepseekmath} and its variants ~\citep{yu2025dapo,liu2025understanding,wang2025beyond}.
However, most of these studies are primarily validated on dense models, leaving the unique challenges of applying RL to MoE models underexplored.

Our observations reveal that directly adapting GRPO-style algorithms to large-scale MoE models leads to severe training instability. The fundamental issue arises from the computational discrepancy between the inference engine and the training engine. In our training framework, the kernels used in these two engines are different, which leads to slight numerical differences. In large-scale MoE models, the use of dynamic expert routing and FP8 quantization significantly amplifies this discrepancy, causing a mismatch between the experts activated during inference and training. As a result, the policies used during inference and training diverge significantly, making the process more off-policy than intended. Furthermore, the introduction of mini-batch updates for accelerating the policy update frequency exacerbates the activated experts' difference between the old and new policies.
Similar issues have been observed in contemporaneous work such as MiniMax-M1~\citep{chen2025minimax} and GSPO~\citep{zheng2025group}. MiniMax-M1 proposes replacing the token-level clipping importance sampling with importance weights to prevent the gradient update of critical tokens from being ignored. On the other hand, GSPO suggests replacing token-level importance sampling with sequence-level importance sampling. Despite their different formulations, both approaches converge on the same insight: token-level clipping based on the ratio of new and old policy log-probabilities is unreliable for MoE models due to the differences in expert routing.

In our work, we adopt our previously proposed OREAL~\citep{lyu2025exploring} algorithm. It utilizes supervised fine-tuning loss for behavior cloning on positive samples and policy gradient on negative samples. It does not introduce token-level clipping based on the ratio of log probabilities between the old and new policies, thereby inherently avoiding the problem of MoE training collapse. However, applying OREAL to large-scale MoE models presents new challenges. Specifically, OREAL requires the online training of a token-level reward model for credit assignment, which requires much higher computational overhead compared to methods like GRPO. To accelerate training, we removed this token-level reward model. Unfortunately, the absence of credit assignment coefficients leads to a rapid reduction in entropy during training, causing the policy model to quickly lose its exploratory capability and converge to suboptimal, thus hindering performance improvement. To mitigate this issue, we incorporate insights from the policy drift in CPGD~\citep{liu2025cpgd} and the KL-Cov strategy from recent entropy mechanism~\citep{cui2025entropy} research. Specifically, we augment the reinforcement learning loss function with a KL divergence constraint term \ref{eq: kl cov}.

\begin{equation}\label{eq: kl cov}
\mathcal{L}_{\text{KL-Cov}}(\theta) =
\begin{cases}
0, & t \notin I, \\[1.2em]
\mathbb{E}_t \left[
- \beta D_{\text{KL}}\!\left( \pi_{\theta_{\text{old}}}(y_t \mid y_{<t}) \,\|\, \pi_\theta(y_t \mid y_{<t}) \right)
\right], & t \in I.
\end{cases}
\end{equation}

This constraint is selectively applied only to tokens whose covariance falls within a specified range $I = \left\{ i \,\middle|\, \mathrm{Rank}\!\left( \mathrm{Cov}(y_i) \right) \le k \cdot N \right\}$ as mentioned in \citet{cui2025entropy}, effectively preventing entropy collapse while maintaining the model's exploration capabilities throughout the training process.
Thus, the overall objective is as follows: 

\begin{equation}
\begin{aligned}\label{eq: policy loss}
\mathcal{L}(\theta)=
\lambda_{\text{sft}}\,\mathbb{E}_{\mathcal{D}^+}\!\big[L_{\text{sft}}(x,y;\theta)\big]
+\lambda_{\text{pg}}\,\mathbb{E}_{\mathcal{D}^-}\!\big[L_{\text{pg}}(x,y;\theta)\big]
+\mathcal{L}_{\text{KL-Cov}}(\theta)
\end{aligned}
\end{equation}

which contains an SFT loss $L_{\text{sft}}$ applied to positive samples $(x,y) \in \mathcal{D}^+$, a policy gradient loss $L_{\text{pg}}$ applied to negative samples $(x,y) \in \mathcal{D}^-$ with an advantage estimate $A(x,y)$, and an entropy control term $\mathcal{L}_{\text{KL-Cov}}(\theta)$. The weighting coefficients $\lambda_{\text{sft}}$ and $\lambda_{\text{pg}}$ balance the contributions of the SFT and policy losses, respectively.

\subsubsection{Experiments}

\paragraph{Hybrid data filtering strategy}

Recent works \citet{he2025skywork,team2025kimi,yu2025dapo} have explored various online and offline filtering methods for reasoning data. Drawing from these approaches, we implement a hybrid strategy that combines offline and online filtering to refine our training data.

For the offline filtering phase, we first perform rollouts on our raw dataset using both a smaller dense SFT model and a large-scale MoE SFT model. Each question is generated 8 times, and the answers are compared against reference solutions for calculating pass rates $\widehat{r}$. Let 

$$
p_{\text{dense}}(x) \in \{0,1\}^{8},\quad
p_{\text{MoE}}(x) \in \{0,1\}^{8},\quad
\widehat{r}_{\text{dense}}(x)=\tfrac{1}{8}\sum p_{\text{dense}}(x),\ 
\widehat{r}_{\text{MoE}}(x)=\tfrac{1}{8}\sum p_{\text{MoE}}(x).
$$

First, we discard items that $\widehat{r}_{\text{dense}}(x)=1.0$, as these questions are deemed excessively simple. Conversely, we also exclude problems that $\widehat{r}_{\text{dense}}(x) \leq 0.25$, which often contain noisy data such as ambiguous questions or mislabeled answers.

In the online filtering phase, each question is also processed in groups of 8 rollouts. We filter out trajectories where all rollouts in a group are either completely correct or completely incorrect. Additionally, among the incorrect samples, we remove data containing garbled text or infinite repetitions. Based on empirical evidence, optimizing policies on such problematic data frequently leads to training collapse.

\begin{figure}[htbp]
    \centering
    \includegraphics[width=0.7\linewidth]{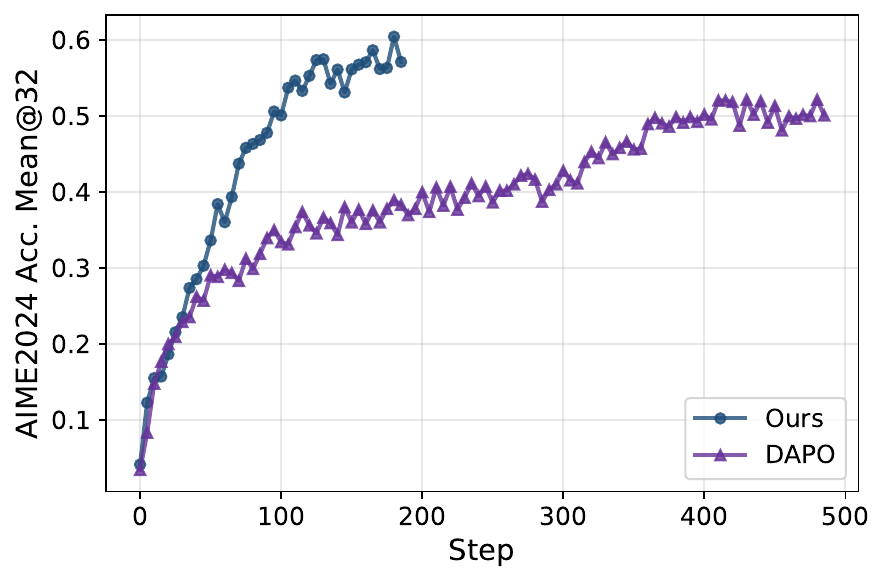}
    \caption{Comparison of 32 times mean accuracy on AIME2024 evaluation set between the model with our strategy and DAPO across training steps for Qwen2.5 32B Base model.}
    \label{fig:math_zero_rl}
\end{figure}

To validate the effectiveness of our data filtering strategy, we perform experiments on the Qwen2.5 32B Base model with mathematical domain data and compare it with DAPO. As shown in Fig~\ref{fig:math_zero_rl}, with our data filtering strategy, the model achieves a significantly faster improvement on the AIME2024 evaluation set compared to DAPO's filtering methodology. Thus, we apply this strategy to all verifiable training data.

\paragraph{Entropy control}

\begin{figure}[htbp]
    \centering
    \begin{subfigure}[b]{0.48\textwidth}
        \centering
        \includegraphics[width=\textwidth]{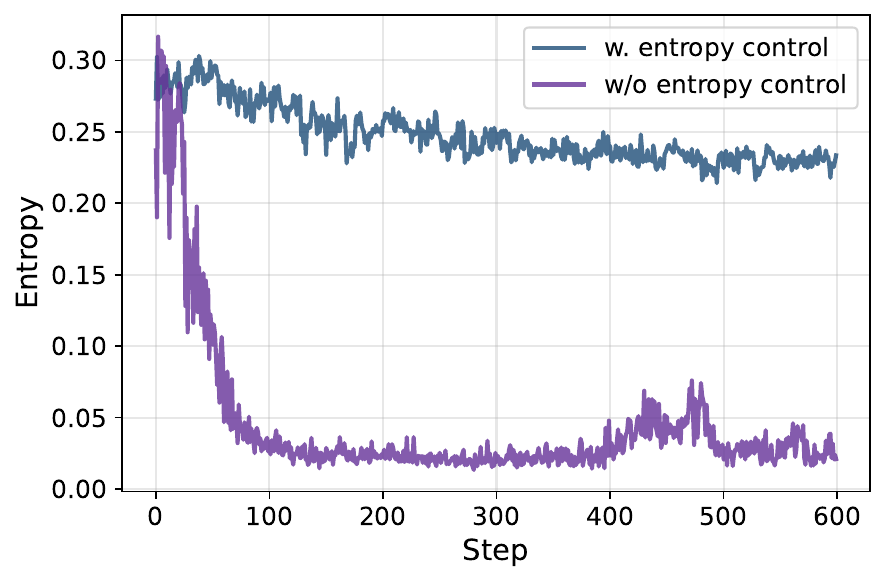}
        \label{fig:entropy_sub1}
    \end{subfigure}
    \hfill
    \begin{subfigure}[b]{0.48\textwidth}
        \centering
        \includegraphics[width=\textwidth]{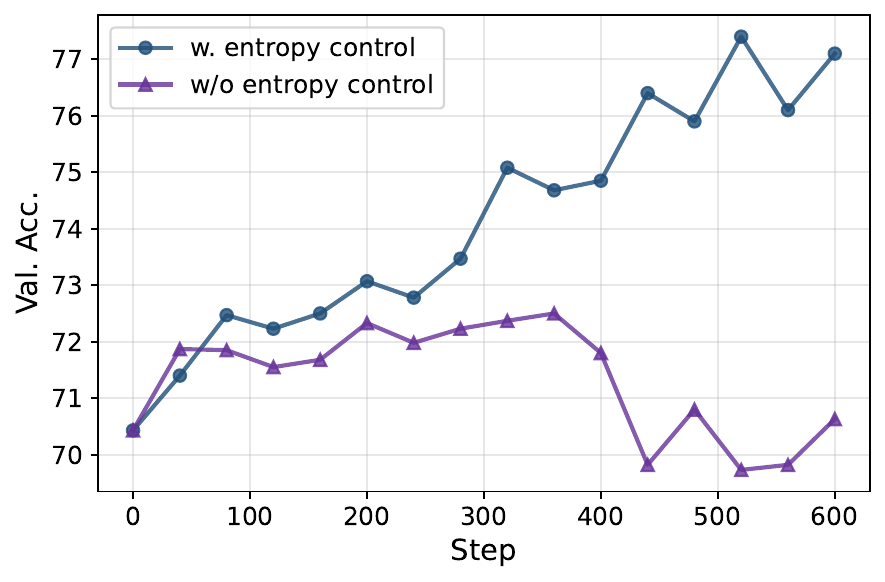}
        \label{fig:entropy_sub2}
    \end{subfigure}
    \caption{Entropy and average validation sets accuracy for Intern-S1 MoE model with and without entropy control during training.}
    \label{fig:entropy_comparison}
\end{figure}

Entropy is a critical factor influencing model exploration and exploitation during RL training. While DAPO proposes the clip-higher mechanism to prevent premature entropy reduction, this approach relies on token-level clipping that, as discussed earlier, cannot be effectively applied to MoE training due to expert activation inconsistencies. We address this limitation by utilizing the KL-Cov entropy control strategy, which maintains exploration throughout the training process. However, the original hyperparameters from \citet{cui2025entropy} were designed for Qwen2.5 model family with high initial entropy. In our case, the Intern-S1 MoE model after cold start exhibited a relatively low initial entropy. To address this, we increased the coefficient of the entropy control term to enhance its effect. In practice, we set the effect token ratio $k$ to 0.2 and the kl coefficient $\beta$ to 0.01. As shown in Fig~\ref{fig:entropy_comparison}, with the entropy control strategy, the model's entropy was maintained at approximately 0.2, and the correctness rate continued to rise across multiple evaluation sets, demonstrating the effectiveness of the entropy control mechanism.

\paragraph{Training details}
For the RL training of the Intern-S1 MoE model, we utilized the data mentioned in Section~\ref{sec: tasks}. When training on open-ended dialogues, we also adopt the chain-of-thought approach for output generation, where only the content after the ``think" segment is submitted to the reward model for scoring. To make the objective compatible with our loss and with other RLVR-style data, we treat samples with $\hat A(x,y)>0$ as positives and the remainder as negatives.

Throughout the RL process, we employ FP8 quantization during both rollout and training phases to enhance computational efficiency. For each prompt, we generate 8 rollout responses, and the training batch size was set to 4096, which is divided into 8 mini-batch steps for updates. We use the AdamW optimizer with a learning rate of 5e-7, weight decay of 0.1, and beta parameters set to 0.9 and 0.95 respectively. During training, the parameters of the ViT encoder and MoE router are frozen. The model is trained for 600 steps in total. To mitigate the effect of noisy samples on stability, we apply a gradient-norm filter: batches with $\text{grad\_norm} > 0.3$ are excluded from gradient updates. Approximately 3\% of training samples are dropped under this criterion. Finally, we select several checkpoints that exhibited the best results on the full evaluation set and perform weight averaging across these checkpoints to achieve more balanced performance.

\section{Evaluation}

We conduct extensive experiments on a variety of benchmarks, including both text-only and multi-modal tasks. Our evaluations can be broadly categorized into two domains: general reasoning and scientific reasoning. In this section, we first introduce the evaluation configuration, followed by a brief description of the benchmarks used. We then present a performance comparison between Intern-S1 and other state-of-the-art models.

\subsection{Evaluation Configuration}
We evaluate models using VLMEvalKit~\citep{duan2024vlmevalkit} and OpenCompass~\citep{2023opencompass}. Unless otherwise specified, we enable the thinking mode (\texttt{enable\_thinking=True}). To mitigate repetition associated with greedy decoding, we adopt sampling decoding strategy; specifically, we set the temperature to 0.7 for Intern-S1 and 0.8 for Intern-S1-mini.

\begin{table}[htbp]
  \centering
  \resizebox{0.55\textwidth}{!}{%
    \begin{tabular}{lll}
      \toprule
      Parameters & Intern-S1 & Intern-S1-mini \\
      \midrule
      \texttt{max\_tokens} & \texttt{65536} & \texttt{65536} \\
      \texttt{temperature} & \texttt{0.7} & \texttt{0.8} \\
      \texttt{top\_p} & \texttt{0.95} & \texttt{0.95} \\
      \texttt{top\_k} & \texttt{50}& \texttt{50} \\
      \texttt{repetition\_penalty} & \texttt{1.0}& \texttt{1.0} \\
      \bottomrule
    \end{tabular}%
  }
  \caption{Decoding parameters used during the evaluation}
  \label{tab:gen-params}
\end{table}

\subsection{Benchmarks}
To evaluate Intern-S1, we select a set of influential, representative benchmarks spanning general and scientific reasoning. 
We select multiple recent scientific-reasoning benchmarks covering the core disciplines of mathematic, physics, chemistry, life science, materials science, and earth science. By encompassing both text-only and multimodal settings, these benchmarks allow fair comparison of professional reasoning across top-tier models. This subsection provides brief introductions to each benchmark.

\subsubsection{General Reasoning}
\paragraph{MMLU-Pro} extends the original MMLU by emphasizing reasoning over recall: it expands answer choices to ten, removes noisy items, and scales to over 12K questions spanning 14 domains, yielding a harder and more robust benchmark for model comparison~\citep{wang2024mmlu}.

\paragraph{GPQA} is a set of 448 expert-written, graduate-level multiple-choice questions in biology, physics, and chemistry designed to be “Google-proof,” meaning non-experts with unrestricted web access still struggle—making it suitable for evaluating deep knowledge and reasoning~\citep{rein2024gpqa}. Our evaluation is conducted on the Diamond subset.

\paragraph{AIME-2025} comprises 30 short-answer problems (AIME I \& II, 2025) from the American Invitational Mathematics Examination, where each answer is an integer from 000–999; it is widely used to probe mathematical reasoning on small, high-difficulty test sets, we use the collected version from OpenCompass~\citep{2023opencompass}.

\paragraph{IFEval} evaluates instruction following via verifiable constraints (\textit{e.g.}, word counts, keyword mentions, formatting), cataloging 25 instruction types across roughly 500 prompts to enable objective, reproducible automatic scoring~\citep{zhou2023instruction}.

\paragraph{MathVista}  targets mathematical reasoning in visual contexts, aggregating 6{,}141 examples from 28 existing datasets and 3 newly created ones (IQTest, FunctionQA, PaperQA), requiring fine-grained visual understanding and compositional reasoning~\citep{lu2023mathvista}.

\paragraph{MMMU}  is a 11.5K-question, college-level multimodal benchmark spanning six core disciplines, 30 subjects, 183 subfields, and 30 image types, designed to test advanced perception and domain-specific reasoning~\citep{yue2024mmmu}.

\paragraph{MathVision} curates 3{,}040 competition-style math problems with visual contexts across 16 disciplines and five difficulty levels, exposing significant gaps between current LMMs and human performance in multimodal math reasoning~\citep{wang2024measuring}.

\paragraph{MMStar} is a vision-indispensable benchmark of 1{,}500 human-curated items balanced over six core capabilities and 18 axes, explicitly filtering out samples solvable without images and proposing metrics to quantify leakage and true multimodal gains~\citep{chen2024we}.

\subsubsection{Scientific Reasoning}

\paragraph{SmolInstruct} is a large-scale instruction-tuning dataset for chemistry centered on small molecules, covering 14 carefully selected tasks and \(\sim\)3.3M query–response pairs, with both SMILES and SELFIES variants \citep{yu2024llasmol}. We use the official test split for evaluation.

\paragraph{ChemBench} is an automated evaluation framework with 2{,}788 curated question–answer pairs spanning undergraduate and graduate chemistry, designed to assess both knowledge and reasoning of LLMs against expert chemists \citep{mirza2025framework}.

\paragraph{MatBench} is a standardized test suite of 13 supervised materials-property prediction tasks drawn from 10 open datasets, with fixed splits and an Automatminer reference baseline to enable fair, repeatable benchmarking \citep{dunn2020benchmarking}.

\paragraph{ProteinLMBench} contains 944 manually verified multiple-choice questions covering sequence, structure, and function, providing a focused benchmark (with an accompanying pretraining/SFT corpus) for evaluating protein understanding in LLMs \citep{shen2024fine}.

\paragraph{SFE}
Scientists’ First Exam (SFE) probes scientific cognition of MLLMs across perception, attribute understanding, and comparative reasoning, comprising 830 expert-verified VQA items across 66 tasks in five disciplines \citep{zhou2025scientists}.

\paragraph{PHYSICS}
benchmarks foundation models on 1{,}297 PhD-qualifying-exam physics problems (including 298 multimodal items) across six core subfields, with automated symbolic-equivalence checking to verify answers \citep{fengphysics}.

\paragraph{MicroVQA} is a microscopy-centric multimodal benchmark with 1{,}042 multiple-choice questions spanning diverse imaging modalities and biological topics, targeting scientific analysis and reasoning \citep{burgess2025microvqa}.

\paragraph{MSEarth-MCQ} is the multiple-choice subset of the MSEarth Earth-science benchmark, providing \(\approx\)2.78K expert-derived figure-grounded questions across atmosphere, cryosphere, hydrosphere, lithosphere, and biosphere \citep{zhao2025msearth,msearthmcq2025hf}.

\paragraph{XLRS-Bench} evaluates MLLMs on ultra-high-resolution remote-sensing imagery (average \(\sim8.5\text{k}\times8.5\text{k}\) pixels) with human-verified annotations across diverse tasks reflecting real-world RS scenarios \citep{wang2025xlrsbenchmultimodalllmsunderstand}.

\subsection{Performance Comparison}
In this section, we compare the Intern-S1 with recently published LLMs and VLMs, including the proprietary API models and open-source models.
\subsubsection{Intern-S1}

\begin{table}[t]
\centering 
\resizebox{\textwidth}{!}{
\tablestyle{2pt}{1.5}
    \begin{tabular}{c|cccc|cccc}
    \shline
    \textbf{Models\textbackslash Mode}    & \multicolumn{4}{c|}{\textbf{Text-Only}}                  & \multicolumn{4}{c}{\textbf{Multi-Modal}}  \\ \shline
\textbf{Benchmarks}       & \textbf{MMLU-Pro} & \textbf{GPQA} & \textbf{AIME2025} & \textbf{IFEval} & \textbf{MathVista} & \textbf{MMMU} & \textbf{MathVision} & \textbf{MMStar} \\  \shline
\rowcolor{gray!15}\multicolumn{9}{c}{\emph{Proprietary API Models}}  \\ 
\textbf{Gemini-2.5 Pro}   & \underline{86.0} & 83.8 & 83.0 & 91.5 & 80.3 & \underline{81.9} & \underline{73.0} & \underline{79.3} \\
\textbf{OpenAI o3} & 85.0 & 83.3 & 88.9 & 92.2 & 77.5 & 80.8 & 67.7  & 75.1 \\
\textbf{Grok-4} & 85.9 & \underline{87.5} & \underline{91.7} & \underline{92.8} & 72.5 & 77.9 & 67.3 & 69.6 \\  
\rowcolor{t_green}\multicolumn{9}{c}{\emph{OpenSource Large Language Models}}  \\ 
\textbf{Deepseek-R1-0528}  & 83.4 & \textbf{80.6 } & \textbf{87.5} & 79.7 & - & - & - & - \\
\textbf{Qwen3-235B-A22B}  & 82.2 & 71.1 & 81.5 & 85.0 & - & - & - & - \\
\textbf{Kimi-K2-Instruct} & 82.7 & 77.8 & 51.4 & \textbf{90.2} & - & - & - & - \\
\rowcolor{orange!30}\multicolumn{9}{c}{\emph{OpenSource Large Multi-Modal Models}}  \\ 
\textbf{InternVL3-78B} & 73.0 & 49.9 & 10.7 & 75.6 & 79.0 & 72.2 & 43.1 & 72.5 \\
\textbf{Qwen2.5-VL-72B} & 72.1 & 49.0 & 10.9 & 83.9 & 74.8 & 70.2 & 38.1 & 70.8 \\  
\textbf{Intern-S1} & \textbf{83.5} & 77.3 & 86.0 & 86.7 & \underline{\textbf{81.5}} & \textbf{77.7} & \textbf{62.5} & \textbf{74.9} \\  \shline
    \end{tabular}}
    \caption{\textbf{The Quantitative Performance of Intern-S1 on General Reasoning Benchmarks. } \underline{Underline} indicates the best performance among all models; \textbf{Bold} indicates the best performance among all open-source models. }
    \label{tab:general:eval}
\end{table}

\begin{table}[t]
\centering 
\resizebox{\textwidth}{!}{
\tablestyle{10pt}{1.5}
    \begin{tabular}{c|cccc}
    \shline
\textbf{Models \textbackslash Benchmarks} & \textbf{SmolInstruct} & \textbf{ChemBench} & \textbf{MatBench} & \textbf{ProteinLMBench} \\  \shline
\rowcolor{gray!15}\multicolumn{5}{c}{\emph{Proprietary API Models}}  \\ 
\textbf{Gemini-2.5 Pro} & 40.4 & 82.8 & 61.7 & 62.9 \\
\textbf{OpenAI o3} & 43.9 & 81.6 & 61.6 & \underline{67.7} \\
\textbf{Grok-4} & 47.3 & 83.3 & 67.9 & 66.2 \\  
\rowcolor{t_green}\multicolumn{5}{c}{\emph{OpenSource Large Language Models}}  \\ 
\textbf{Deepseek-R1-0528}  & 30.7 & 75.6 & 57.7 & 61.4 \\
\textbf{Qwen3-235B-A22B}  & 28.7 & 75.8 & 52.1 & 59.8 \\
\textbf{Kimi-K2-Instruct} & 48.1 & 75.3 & 61.7 & \textbf{66.7} \\
\rowcolor{orange!30}\multicolumn{5}{c}{\emph{OpenSource Large Multi-Modal Models}}  \\ 
\textbf{InternVL3-78B} & 19.4 & 61.3 & 49.3 & 61.6 \\
\textbf{Qwen2.5-VL-72B} & 21.0 & 61.6 & 51.5 & 61.0 \\  
\textbf{Intern-S1} & \underline{\textbf{51.0}} & \underline{\textbf{83.4}} & \underline{\textbf{75.0}} & 63.1 \\ \shline
    \end{tabular}}
    \caption{\textbf{The Quantitative Performance of Intern-S1 on Science-related Benchmarks (Text-Only). } \underline{Underline} indicates the best performance among all models; \textbf{Bold} indicates the best performance among all open-source models. }
    \label{tab:science:text:eval}
\end{table}

\begin{table}[t]
\centering 
\resizebox{\textwidth}{!}{
\tablestyle{8pt}{1.5}
    \begin{tabular}{c|ccccc}
    \shline
\textbf{Models \textbackslash Benchmarks} & \textbf{SFE} & \textbf{Physics} & \textbf{MicroVQA} & \textbf{MSEarthMCQ} & \textbf{XLRS-Bench} \\  \shline
\rowcolor{gray!15}\multicolumn{6}{c}{\emph{Proprietary API Models}}  \\ 
\textbf{Gemini-2.5 Pro} & 43.0 & 40.0 & 63.1 & 59.9 & 45.2 \\
\textbf{OpenAI o3} & 37.7 & \underline{47.9} & 58.3 & 61.0 & 43.6 \\
\textbf{Grok-4} & 31.2 & 42.8 & 59.5 & 58.0 & 45.4\\  
\rowcolor{orange!30}\multicolumn{6}{c}{\emph{OpenSource Large Multi-Modal Models}}  \\ 
\textbf{InternVL3-78B} & 36.2 & 23.1 & 59.1 & 57.2 & 49.3 \\
\textbf{Qwen2.5-VL-72B} & 30.5 & 15.7 & 53.0 & 37.6 & 50.9 \\  
\textbf{Intern-S1} & \underline{\textbf{44.3}} & \textbf{44.0} & \underline{\textbf{63.9}} & \underline{\textbf{65.7}} & \underline{\textbf{55.0}} \\ \shline
    \end{tabular}}
    \caption{\textbf{The Quantitative Performance of Intern-S1 on Science-related Benchmarks (Multi-Modal). } \underline{Underline} indicates the best performance among all models; \textbf{Bold} indicates the best performance among all open-source models. }
    \label{tab:science:mm:eval}
\end{table}
Intern-S1 delivers state-of-the-art open-source performance and is competitive with proprietary APIs, especially on the scientific reasoning.

On general-reasoning benchmarks(Tab.~\ref{tab:general:eval}) , Intern-S1 is the top open-source multimodal model across all eight tasks, and it achieves the best overall result on MathVista (81.5). It substantially outperforms prior open-source MLLMs—\textit{e.g.}, +2.5/+6.7 on MathVista vs. InternVL3-78B/Qwen2.5-VL-72B, and +19.4/+24.4 on MathVision—though it still trails leading API models on text-only GPQA and IFEval, and is competitive on AIME2025.

On science-focused text-only benchmarks(Tab.~\ref{tab:science:text:eval}), Intern-S1 takes best overall on three of four datasets—SmolInstruct (51.0), ChemBench (83.4), and MatBench (75.0)—with large margins over previous open-source MLLMs (\textit{e.g.}, MatBench: +25.7 vs. InternVL3-78B, +23.5 vs. Qwen2.5-VL-72B). On science-focused multimodal benchmarks(Tab.~\ref{tab:science:mm:eval}), Intern-S1 attains the best overall scores on 4/5 datasets—SFE (44.3), MicroVQA (63.9), MSEarthMCQ (65.7), and XLRS-Bench (55.0)—and ranks second on Physics (44.0 vs. 47.9 for o3). Gains over open-source baselines are consistent and sizable.

Overall, the results indicate that Intern-S1 markedly narrows the gap to close-source large models in general reasoning and sets a new bar for open-source multimodal scientific reasoning, with remaining headroom on instruction-following constraints.

\begin{table}[t]
\centering 
\resizebox{\textwidth}{!}{
\tablestyle{2pt}{1.5}
    \begin{tabular}{c|cccc|cccc}
    \shline
    \textbf{Models\textbackslash Mode}    & \multicolumn{4}{c|}{\textbf{Text-Only}}                  & \multicolumn{4}{c}{\textbf{Multi-Modal}}  \\ \shline
\textbf{Benchmarks}       & \textbf{MMLU-Pro} & \textbf{GPQA} & \textbf{AIME2025} & \textbf{IFEval} & \textbf{MathVista} & \textbf{MMMU} & \textbf{MathVision} & \textbf{MMStar} \\  \shline
  
\rowcolor{t_green}\multicolumn{9}{c}{\emph{OpenSource Large Language Models}}  \\ 
\textbf{Qwen3-8B}  & 73.7 & 62.0 & 67.3 & \textbf{85.0} & - & - & - & - \\
\rowcolor{orange!30}\multicolumn{9}{c}{\emph{OpenSource Large Multi-Modal Models}}  \\ 
\textbf{GLM-4.1V-Thinking} & 57.1 & 50.3 & 32.0 & 71.5 & \textbf{80.7} & 69.9 & \textbf{53.9} & 71.5 \\
\textbf{MiMo-VL-7B-RL-2508} & 73.9 & 60.4 & 64.4 & 71.4 & 79.4 & 70.6 & 38.1 & \textbf{72.9} \\  
    \textbf{Intern-S1-mini} & \textbf{74.8} & \textbf{65.2} & \textbf{80.0} & 81.2 & 70.3 & \textbf{72.3}   & 51.4 & 65.2 \\  \shline
    \end{tabular}}
    \caption{\textbf{The Quantitative Performance of Intern-S1-mini on General Reasoning Benchmarks. } 
    \textbf{Bold} indicates the best performance among all open-source models. }
    \label{tab:general:mini_eval}
\end{table}

\begin{table}[t]
\centering 
\resizebox{\textwidth}{!}{
\tablestyle{10pt}{1.5}
    \begin{tabular}{c|cccc}
    \shline
\textbf{Models \textbackslash Benchmarks} & \textbf{SmolInstruct} & \textbf{ChemBench} & \textbf{MatBench} & \textbf{ProteinLMBench} \\  \shline
\rowcolor{t_green}\multicolumn{5}{c}{\emph{OpenSource Large Language Models}}  \\ 
\textbf{Qwen3-8B} & 17.6 & 61.1 & 45.2 & 59.1 \\
\rowcolor{orange!30}\multicolumn{5}{c}{\emph{OpenSource Large Multi-Modal Models}}  \\ 
\textbf{GLM-4.1V-Thinking} & 18.1 & 56.2 & 54.3 & 58.3 \\
\textbf{MiMo-VL-7B-RL-2508} & 16.1 & 66.8 & 46.9 & \textbf{59.8} \\  
\textbf{Intern-S1-mini} & \textbf{32.2} & \textbf{76.5} &  \textbf{61.6} & 63.1 \\ \shline
    \end{tabular}}
    \caption{\textbf{The Quantitative Performance of Intern-S1-mini on Science-related Benchmarks (Text-Only). } 
    \textbf{Bold} indicates the best performance among all open-source models. }
    \label{tab:science:text:mini_eval}
\end{table}

\begin{table}[t]
\centering 
\resizebox{\textwidth}{!}{
\tablestyle{8pt}{1.5}
    \begin{tabular}{c|ccccc}
    \shline
\textbf{Models \textbackslash Benchmarks} & \textbf{SFE} & \textbf{Physics} & \textbf{MicroVQA} & \textbf{MSEarthMCQ} & \textbf{XLRS-Bench} \\  \shline
\rowcolor{orange!30}\multicolumn{6}{c}{\emph{OpenSource Large Multi-Modal Models}}  \\ 
\textbf{GLM-4.1V-Thinking} & 43.2 & 28.3 & 50.2 & 50.3 & 49.8 \\
\textbf{MiMo-VL-7B-RL-2508} & \textbf{43.9} & 28.2 & 51.0 & 47.3 & 12.3 \\  
\textbf{Intern-S1-mini} & 35.8 & \textbf{28.8} & \textbf{56.6} & \textbf{58.1} & \textbf{51.6} \\ \shline
    \end{tabular}}
    \caption{\textbf{The Quantitative Performance of Intern-S1-mini on Science-related Benchmarks (Multi-Modal). } \textbf{Bold} indicates the best performance among all open-source models. }
    \label{tab:science:mm:mini_eval}
\end{table}

\subsubsection{Intern-S1-mini}
We further conduct comparison on Intern-S1-mini with the recently open-sourced LLMs and VLMs, including the Qwen3-8B~\citep{yang2025qwen3}, GLM-4.1V-Thinking~\citep{hong2025glm} and MiMo-VL-7B-RL-2508~\citep{xiaomi2025mimo}. Results are summarized in Table.~\ref{tab:general:mini_eval},~\ref{tab:science:text:mini_eval},~\ref{tab:science:mm:mini_eval}.

\noindent\textbf{General reasoning (text-only).}  
As shown in Table~\ref{tab:general:mini_eval}, Intern-S1-mini sets a new open-source state of the art across multiple text-only benchmarks: \textbf{74.8} on MMLU-Pro (\(+0.9\) over the best baseline, 73.9), \textbf{65.2} on GPQA (\(+3.2\) over 62.0), and \textbf{80.0} on AIME2025 (\(+12.7\) over 67.3). These results indicate strong factual reasoning, competition-level math ability, and instruction-following.

\noindent\textbf{General reasoning (multi-modal).}  
On the visual reasoning side (Table~\ref{tab:general:mini_eval}), Intern-S1-mini achieves the best open-source score on \textbf{MMMU} with \textbf{72.3} (\(+1.7\) over 70.6). Performance on \textit{MathVista} (70.3) and \textit{MathVision} (51.4) is competitive but trails GLM-4.1V-Thinking, and on \textit{MMStar} (65.2) is \(7.7\) behind MiMo-VL-7B-RL-2508. These gaps suggest remaining headroom on visually intensive math problems.

\noindent\textbf{Scientific reasoning (text-only).}  
Table~\ref{tab:science:text:mini_eval} shows substantial gains on domain scientific benchmarks. Intern-S1-mini attains \textbf{32.2} on SmolInstruct (\(+14.1\)), \textbf{76.5} on ChemBench (\(+9.7\)), \textbf{75.0} on MatBench (\(+20.7\)), and \textbf{63.1} on ProteinLMBench (\(+3.3\)) over the strongest open-source baselines. The consistent margins indicate strong scientific knowledge and compositional reasoning.

\noindent\textbf{Scientific reasoning (multi-modal).}  
On the multi-modal scientific suite (Table~\ref{tab:science:mm:mini_eval}), Intern-S1-mini delivers the best scores on \textbf{4/5} datasets: Physics \textbf{28.8} (\(+4.9\) over 23.9), MicroVQA \textbf{56.6} (\(+5.6\)), MSEarthMCQ \textbf{58.1} (\(+7.8\)), and XLRS-Bench \textbf{51.6} (\(+1.8\)).

Overall, across 17 benchmarks, \textbf{Intern-S1-mini} achieves the top open-source score on \textbf{11} of them, matching or exceeding the strong text-only LLM \emph{Qwen3-8B} and remaining highly competitive with contemporary VLMs. These results indicate that Intern-S1-mini is not only a competitive general-purpose VLM but also a superior \emph{lightweight} scientific reasoner. We expect this powerful multimodal model to facilitate a broad range of science-related tasks.

 \section{Conclusion}
 In this report, we have introduced Intern-S1, the state-of-the-art scientific multi-modal model, and its lightweight version, Intern-S1-mini. Intern-S1 has the top-tier general reasoning ability among open-source models and the top-tier scientific reasoning ability compared to closed-source models. With over 2.5T of scientific data in the pre-training, Intern-S1 has a broad knowledge background related to scientific tasks, ensuring it can serve as a good foundation model for scientific research and applications. By leveraging the InternBootCamp and Mixture-of-Rewards frameworks, our RL training process achieves a cost reduction of 10X compared to publicly available baselines, and the high sample efficiency is important to teach models the professional skills in scientific domains.

\subsubsection*{Author Contributions}
The authors are listed in alphabetical order by their last names.

\paragraph{Contributors:} Lei Bai, Zhongrui Cai, Yuhang Cao, Maosong Cao, Weihan Cao, Chiyu Chen, Haojiong Chen, Kai Chen, Pengcheng Chen, Ying Chen, Yongkang Chen, Yu Cheng, Pei Chu, Tao Chu, Erfei Cui, Ganqu Cui, Long Cui, Ziyun Cui, Nianchen Deng, Ning Ding, Nanqing Dong, Peijie Dong, Shihan Dou, Sinan Du, Haodong Duan, Caihua Fan, Ben Gao, Changjiang Gao, Jianfei Gao, Songyang Gao, Yang Gao, Zhangwei Gao, Jiaye Ge, Qiming Ge, Lixin Gu, Yuzhe Gu, Aijia Guo, Qipeng Guo, Xu Guo, Conghui He, Junjun He, Yili Hong, Siyuan Hou, Caiyu Hu , Hanglei Hu, Jucheng Hu, Ming Hu, Zhouqi Hua, Haian Huang, Junhao Huang, Xu Huang, Zixian Huang, Zhe Jiang, Lingkai Kong, Linyang Li , Peiji Li , Pengze Li, Shuaibin Li, Tianbin Li, Wei Li, Yuqiang Li, Dahua Lin, Junyao Lin, Tianyi Lin, Zhishan Lin, Hongwei Liu, Jiangning Liu, Jiyao Liu, Junnan Liu, Kai Liu, Kaiwen Liu, Kuikun Liu, Shichun Liu, Shudong Liu, Wei Liu, Xinyao Liu, Yuhong Liu, Zhan Liu, Yinquan Lu, Haijun Lv, Hongxia Lv, Huijie Lv, Qitan Lv, Ying Lv, Chengqi Lyu, Chenglong Ma, Jianpeng Ma, Ren Ma, Runmin Ma, Runyuan Ma, Xinzhu Ma, Yichuan Ma, Zihan Ma, Sixuan Mi, Junzhi Ning, Wenchang Ning, Xinle Pang, Jiahui Peng, Runyu Peng, Yu Qiao, Jiantao Qiu, Xiaoye Qu, Yuan Qu, Yuchen Ren, Fukai Shang, Wenqi Shao, Junhao Shen, Shuaike Shen, Chunfeng Song, Demin Song, Diping Song, Chenlin Su, Weijie Su, Weigao Sun, Yu Sun, Qian Tan, Cheng Tang, Huanze Tang, Kexian Tang, Shixiang Tang, Jian Tong, Aoran Wang, Bin Wang, Dong Wang, Lintao Wang, Rui Wang, Weiyun Wang, Wenhai Wang, Jiaqi Wang, Yi Wang, Ziyi Wang, Ling-I Wu, Wen Wu, Yue Wu, Zijian Wu, Linchen Xiao, Shuhao Xing , Chao Xu, Huihui Xu, Jun Xu, Ruiliang Xu, Wanghan Xu, GanLin Yang, Yuming Yang, Haochen Ye, Jin Ye, Shenglong Ye, Jia Yu, Jiashuo Yu, Jing Yu, Fei Yuan, Yuhang Zang, Bo Zhang, Chao Zhang, Chen Zhang, Hongjie Zhang, Jin Zhang, Qiaosheng Zhang, Qiuyinzhe Zhang, Songyang Zhang, Taolin Zhang, Wenlong Zhang, Wenwei Zhang, Yechen Zhang, Ziyang Zhang, Haiteng Zhao, Qian Zhao, Xiangyu Zhao, Xiangyu Zhao, Bowen Zhou, Dongzhan Zhou, Peiheng Zhou, Yuhao Zhou, Yunhua Zhou, Dongsheng Zhu, Lin Zhu, Yicheng Zou

\bibliography{iclr2025_conference}

\begin{thebibliography}{80}
\providecommand{\natexlab}[1]{#1}
\providecommand{\url}[1]{\texttt{#1}}
\expandafter\ifx\csname urlstyle\endcsname\relax
  \providecommand{\doi}[1]{doi: #1}\else
  \providecommand{\doi}{doi: \begingroup \urlstyle{rm}\Url}\fi

\bibitem[Anthropic(2025)]{claude4}
Anthropic.
\newblock Introducing claude 4, 2025.
\newblock URL \url{https://www.anthropic.com/news/claude-4}.

\bibitem[Bai et~al.(2025)Bai, Chen, Liu, Wang, Ge, Song, Dang, Wang, Wang, Tang, et~al.]{bai2025qwen2}
Shuai Bai, Keqin Chen, Xuejing Liu, Jialin Wang, Wenbin Ge, Sibo Song, Kai Dang, Peng Wang, Shijie Wang, Jun Tang, et~al.
\newblock Qwen2. 5-vl technical report.
\newblock \emph{arXiv preprint arXiv:2502.13923}, 2025.

\bibitem[Bai et~al.(2022)Bai, Jones, Ndousse, Askell, Chen, DasSarma, Drain, Fort, Ganguli, Henighan, et~al.]{bai2022training}
Yuntao Bai, Andy Jones, Kamal Ndousse, Amanda Askell, Anna Chen, Nova DasSarma, Dawn Drain, Stanislav Fort, Deep Ganguli, Tom Henighan, et~al.
\newblock Training a helpful and harmless assistant with reinforcement learning from human feedback.
\newblock \emph{arXiv preprint arXiv:2204.05862}, 2022.

\bibitem[Bi et~al.(2024)Bi, Chen, Chen, Chen, Dai, Deng, Ding, Dong, Du, Fu, et~al.]{bi2024deepseek}
Xiao Bi, Deli Chen, Guanting Chen, Shanhuang Chen, Damai Dai, Chengqi Deng, Honghui Ding, Kai Dong, Qiushi Du, Zhe Fu, et~al.
\newblock Deepseek llm: Scaling open-source language models with longtermism.
\newblock \emph{arXiv preprint arXiv:2401.02954}, 2024.

\bibitem[Burgess et~al.(2025)Burgess, Nirschl, Bravo-S{\'a}nchez, Lozano, Gupte, Galaz-Montoya, Zhang, Su, Bhowmik, Coman, et~al.]{burgess2025microvqa}
James Burgess, Jeffrey~J Nirschl, Laura Bravo-S{\'a}nchez, Alejandro Lozano, Sanket~Rajan Gupte, Jesus~G Galaz-Montoya, Yuhui Zhang, Yuchang Su, Disha Bhowmik, Zachary Coman, et~al.
\newblock Microvqa: A multimodal reasoning benchmark for microscopy-based scientific research.
\newblock In \emph{Proceedings of the Computer Vision and Pattern Recognition Conference}, pp.\  19552--19564, 2025.

\bibitem[Cao et~al.(2025)Cao, Zhang, Li, Zhang, Liu, Duan, Zhang, and Chen]{cao2025condor}
Maosong Cao, Taolin Zhang, Mo~Li, Chuyu Zhang, Yunxin Liu, Haodong Duan, Songyang Zhang, and Kai Chen.
\newblock Condor: Enhance llm alignment with knowledge-driven data synthesis and refinement.
\newblock \emph{arXiv preprint arXiv:2501.12273}, 2025.

\bibitem[Chang et~al.(2024)Chang, Cui, Dong, Huang, Huang, Huang, Li, Lv, Ma, Sun, et~al.]{chang2024redstone}
Yaoyao Chang, Lei Cui, Li~Dong, Shaohan Huang, Yangyu Huang, Yupan Huang, Scarlett Li, Tengchao Lv, Shuming Ma, Qinzheng Sun, et~al.
\newblock Redstone: Curating general, code, math, and qa data for large language models.
\newblock \emph{arXiv preprint arXiv:2412.03398}, 2024.

\bibitem[Chen et~al.(2025)Chen, Li, Gong, Jiang, Fei, Yang, Shan, Yu, Wang, Zhu, et~al.]{chen2025minimax}
Aili Chen, Aonian Li, Bangwei Gong, Binyang Jiang, Bo~Fei, Bo~Yang, Boji Shan, Changqing Yu, Chao Wang, Cheng Zhu, et~al.
\newblock Minimax-m1: Scaling test-time compute efficiently with lightning attention.
\newblock \emph{arXiv preprint arXiv:2506.13585}, 2025.

\bibitem[Chen et~al.(2024{\natexlab{a}})Chen, Li, Dong, Zhang, Zang, Chen, Duan, Wang, Qiao, Lin, et~al.]{chen2024we}
Lin Chen, Jinsong Li, Xiaoyi Dong, Pan Zhang, Yuhang Zang, Zehui Chen, Haodong Duan, Jiaqi Wang, Yu~Qiao, Dahua Lin, et~al.
\newblock Are we on the right way for evaluating large vision-language models?
\newblock \emph{Advances in Neural Information Processing Systems}, 37:\penalty0 27056--27087, 2024{\natexlab{a}}.

\bibitem[Chen et~al.(2024{\natexlab{b}})Chen, Wang, Tian, Ye, Gao, Cui, Tong, Hu, Luo, Ma, et~al.]{chen2024far}
Zhe Chen, Weiyun Wang, Hao Tian, Shenglong Ye, Zhangwei Gao, Erfei Cui, Wenwen Tong, Kongzhi Hu, Jiapeng Luo, Zheng Ma, et~al.
\newblock How far are we to gpt-4v? closing the gap to commercial multimodal models with open-source suites.
\newblock \emph{Science China Information Sciences}, 67\penalty0 (12):\penalty0 220101, 2024{\natexlab{b}}.

\bibitem[Chen et~al.(2024{\natexlab{c}})Chen, Wu, Wang, Su, Chen, Xing, Zhong, Zhang, Zhu, Lu, et~al.]{chen2024internvl}
Zhe Chen, Jiannan Wu, Wenhai Wang, Weijie Su, Guo Chen, Sen Xing, Muyan Zhong, Qinglong Zhang, Xizhou Zhu, Lewei Lu, et~al.
\newblock Internvl: Scaling up vision foundation models and aligning for generic visual-linguistic tasks.
\newblock In \emph{Proceedings of the IEEE/CVF conference on computer vision and pattern recognition}, pp.\  24185--24198, 2024{\natexlab{c}}.

\bibitem[Contributors(2023{\natexlab{a}})]{2023lmdeploy}
LMDeploy Contributors.
\newblock Lmdeploy: A toolkit for compressing, deploying, and serving llm.
\newblock \url{https://github.com/InternLM/lmdeploy}, 2023{\natexlab{a}}.

\bibitem[Contributors(2023{\natexlab{b}})]{2023opencompass}
OpenCompass Contributors.
\newblock Opencompass: A universal evaluation platform for foundation models.
\newblock \url{https://github.com/open-compass/opencompass}, 2023{\natexlab{b}}.

\bibitem[Contributors(2023{\natexlab{c}})]{2023xtuner}
XTuner Contributors.
\newblock Xtuner: A toolkit for efficiently fine-tuning llm.
\newblock \url{https://github.com/InternLM/xtuner}, 2023{\natexlab{c}}.

\bibitem[Cui et~al.(2023)Cui, Yuan, Ding, Yao, He, Zhu, Ni, Xie, Xie, Lin, et~al.]{cui2023ultrafeedback}
Ganqu Cui, Lifan Yuan, Ning Ding, Guanming Yao, Bingxiang He, Wei Zhu, Yuan Ni, Guotong Xie, Ruobing Xie, Yankai Lin, et~al.
\newblock Ultrafeedback: Boosting language models with scaled ai feedback.
\newblock \emph{arXiv preprint arXiv:2310.01377}, 2023.

\bibitem[Cui et~al.(2025)Cui, Zhang, Chen, Yuan, Wang, Zuo, Li, Fan, Chen, Chen, et~al.]{cui2025entropy}
Ganqu Cui, Yuchen Zhang, Jiacheng Chen, Lifan Yuan, Zhi Wang, Yuxin Zuo, Haozhan Li, Yuchen Fan, Huayu Chen, Weize Chen, et~al.
\newblock The entropy mechanism of reinforcement learning for reasoning language models.
\newblock \emph{arXiv preprint arXiv:2505.22617}, 2025.

\bibitem[DeepMind(2025)]{gemini2.5}
Google DeepMind.
\newblock Gemini 2.5, 2025.
\newblock URL \url{https://blog.google/technology/google-deepmind/gemini-model-thinking-updates-march-2025/}.

\bibitem[Dong et~al.(2025)Dong, Jiang, Tao, Liu, Zhang, Mou, Cao, Ma, Chen, Li, et~al.]{dong2025rl}
Yihong Dong, Xue Jiang, Yongding Tao, Huanyu Liu, Kechi Zhang, Lili Mou, Rongyu Cao, Yingwei Ma, Jue Chen, Binhua Li, et~al.
\newblock Rl-plus: Countering capability boundary collapse of llms in reinforcement learning with hybrid-policy optimization.
\newblock \emph{arXiv preprint arXiv:2508.00222}, 2025.

\bibitem[Dou et~al.(2025)Dou, Liu, Yang, Zou, Zhou, Xing, Huang, Ge, Song, Lv, et~al.]{dou2025pre}
Shihan Dou, Shichun Liu, Yuming Yang, Yicheng Zou, Yunhua Zhou, Shuhao Xing, Chenhao Huang, Qiming Ge, Demin Song, Haijun Lv, et~al.
\newblock Pre-trained policy discriminators are general reward models.
\newblock \emph{arXiv preprint arXiv:2507.05197}, 2025.

\bibitem[Du et~al.(2025)Du, Yao, Ma, Wang, Zheng, Zhu, Liu, Liang, Jin, Wei, et~al.]{du2025supergpqa}
Xinrun Du, Yifan Yao, Kaijing Ma, Bingli Wang, Tianyu Zheng, King Zhu, Minghao Liu, Yiming Liang, Xiaolong Jin, Zhenlin Wei, et~al.
\newblock Supergpqa: Scaling llm evaluation across 285 graduate disciplines.
\newblock \emph{arXiv preprint arXiv:2502.14739}, 2025.

\bibitem[Duan et~al.(2024)Duan, Yang, Qiao, Fang, Chen, Liu, Dong, Zang, Zhang, Wang, et~al.]{duan2024vlmevalkit}
Haodong Duan, Junming Yang, Yuxuan Qiao, Xinyu Fang, Lin Chen, Yuan Liu, Xiaoyi Dong, Yuhang Zang, Pan Zhang, Jiaqi Wang, et~al.
\newblock Vlmevalkit: An open-source toolkit for evaluating large multi-modality models.
\newblock In \emph{Proceedings of the 32nd ACM international conference on multimedia}, pp.\  11198--11201, 2024.

\bibitem[Dunn et~al.(2020)Dunn, Wang, Ganose, Dopp, and Jain]{dunn2020benchmarking}
Alexander Dunn, Qi~Wang, Alex Ganose, Daniel Dopp, and Anubhav Jain.
\newblock Benchmarking materials property prediction methods: the matbench test set and automatminer reference algorithm.
\newblock \emph{npj Computational Materials}, 6\penalty0 (1):\penalty0 138, 2020.

\bibitem[Feher et~al.(2024)Feher, Vuli{\'c}, and Minixhofer]{feher2024retrofitting}
Darius Feher, Ivan Vuli{\'c}, and Benjamin Minixhofer.
\newblock Retrofitting large language models with dynamic tokenization.
\newblock \emph{arXiv preprint arXiv:2411.18553}, 2024.

\bibitem[Feng et~al.()Feng, Zhao, Liu, Yang, Zhao, Sous, and Cohan]{fengphysics}
Kaiyue Feng, Yilun Zhao, Yixin Liu, Tianyu Yang, Chen Zhao, John Sous, and Arman Cohan.
\newblock Physics: Benchmarking foundation models for problem solving in physics.
\newblock In \emph{Workshop on Reasoning and Planning for Large Language Models}.

\bibitem[Ganguli et~al.(2022)Ganguli, Lovitt, Kernion, Askell, Bai, Kadavath, Mann, Perez, Schiefer, Ndousse, et~al.]{ganguli2022red}
Deep Ganguli, Liane Lovitt, Jackson Kernion, Amanda Askell, Yuntao Bai, Saurav Kadavath, Ben Mann, Ethan Perez, Nicholas Schiefer, Kamal Ndousse, et~al.
\newblock Red teaming language models to reduce harms: Methods, scaling behaviors, and lessons learned.
\newblock \emph{arXiv preprint arXiv:2209.07858}, 2022.

\bibitem[Guo et~al.(2025{\natexlab{a}})Guo, Yang, Zhang, Song, Zhang, Xu, Zhu, Ma, Wang, Bi, et~al.]{guo2025deepseek}
Daya Guo, Dejian Yang, Haowei Zhang, Junxiao Song, Ruoyu Zhang, Runxin Xu, Qihao Zhu, Shirong Ma, Peiyi Wang, Xiao Bi, et~al.
\newblock Deepseek-r1: Incentivizing reasoning capability in llms via reinforcement learning.
\newblock \emph{arXiv preprint arXiv:2501.12948}, 2025{\natexlab{a}}.

\bibitem[Guo et~al.(2025{\natexlab{b}})Guo, Liang, Jian, Yang, Wu, Li, Lu, Guo, and Chen]{guo2025ifdecorator}
Xu~Guo, Tianyi Liang, Tong Jian, Xiaogui Yang, Ling-I Wu, Chenhui Li, Zhihui Lu, Qipeng Guo, and Kai Chen.
\newblock Ifdecorator: Wrapping instruction following reinforcement learning with verifiable rewards.
\newblock \emph{arXiv preprint arXiv:2508.04632}, 2025{\natexlab{b}}.

\bibitem[He et~al.(2025)He, Liu, Liu, Yan, Wang, Cheng, Zhang, Zhang, Xu, Shen, Li, Zeng, Wei, Cheng, An, Liu, and Zhou]{he2025skywork}
Jujie He, Jiacai Liu, Chris~Yuhao Liu, Rui Yan, Chaojie Wang, Peng Cheng, Xiaoyu Zhang, Fuxiang Zhang, Jiacheng Xu, Wei Shen, Siyuan Li, Liang Zeng, Tianwen Wei, Cheng Cheng, Bo~An, Yang Liu, and Yahui Zhou.
\newblock Skywork open reasoner 1 technical report.
\newblock \emph{arXiv preprint arXiv:2505.22312}, 2025.

\bibitem[Hong et~al.(2025)Hong, Yu, Gu, Wang, Gan, Tang, Cheng, Qi, Ji, Pan, et~al.]{hong2025glm}
Wenyi Hong, Wenmeng Yu, Xiaotao Gu, Guo Wang, Guobing Gan, Haomiao Tang, Jiale Cheng, Ji~Qi, Junhui Ji, Lihang Pan, et~al.
\newblock Glm-4.1 v-thinking: Towards versatile multimodal reasoning with scalable reinforcement learning.
\newblock \emph{arXiv preprint arXiv:2507.01006}, 2025.

\bibitem[Hsu et~al.(2025)Hsu, Dai, Kothapalli, Song, Tang, Zhu, Shimizu, Sahni, Ning, Chen, and Wang]{hsu2025ligerkernel}
Pin-Lun Hsu, Yun Dai, Vignesh Kothapalli, Qingquan Song, Shao Tang, Siyu Zhu, Steven Shimizu, Shivam Sahni, Haowen Ning, Yanning Chen, and Zhipeng Wang.
\newblock Liger-kernel: Efficient triton kernels for llm training.
\newblock In \emph{Championing Open-source DEvelopment in ML Workshop @ ICML25}, 2025.
\newblock URL \url{https://openreview.net/forum?id=36SjAIT42G}.

\bibitem[Joulin et~al.(2016)Joulin, Grave, Bojanowski, Douze, J{\'e}gou, and Mikolov]{joulin2016fasttext}
Armand Joulin, Edouard Grave, Piotr Bojanowski, Matthijs Douze, H{\'e}rve J{\'e}gou, and Tomas Mikolov.
\newblock Fasttext. zip: Compressing text classification models.
\newblock \emph{arXiv preprint arXiv:1612.03651}, 2016.

\bibitem[Li et~al.(2023)Li, Zhang, Li, Chen, Chen, Cheng, Wang, Zhou, and Xiao]{li2023quantity}
Ming Li, Yong Zhang, Zhitao Li, Jiuhai Chen, Lichang Chen, Ning Cheng, Jianzong Wang, Tianyi Zhou, and Jing Xiao.
\newblock From quantity to quality: Boosting llm performance with self-guided data selection for instruction tuning.
\newblock \emph{arXiv preprint arXiv:2308.12032}, 2023.

\bibitem[Li et~al.(2025)Li, Ye, Chen, Ma, Yu, Chen, Cui, Li, Chen, Lyu, et~al.]{li2025internbootcamp}
Peiji Li, Jiasheng Ye, Yongkang Chen, Yichuan Ma, Zijie Yu, Kedi Chen, Ganqu Cui, Haozhan Li, Jiacheng Chen, Chengqi Lyu, et~al.
\newblock Internbootcamp technical report: Boosting llm reasoning with verifiable task scaling.
\newblock \emph{arXiv preprint arXiv:2508.08636}, 2025.

\bibitem[Liu et~al.(2024)Liu, Feng, Xue, Wang, Wu, Lu, Zhao, Deng, Zhang, Ruan, et~al.]{liu2024deepseek}
Aixin Liu, Bei Feng, Bing Xue, Bingxuan Wang, Bochao Wu, Chengda Lu, Chenggang Zhao, Chengqi Deng, Chenyu Zhang, Chong Ruan, et~al.
\newblock Deepseek-v3 technical report.
\newblock \emph{arXiv preprint arXiv:2412.19437}, 2024.

\bibitem[Liu et~al.(2025{\natexlab{a}})Liu, Liu, Liu, Xiao, Gao, Lyu, Gu, Zhang, Wong, Zhang, et~al.]{liu2025compassverifier}
Shudong Liu, Hongwei Liu, Junnan Liu, Linchen Xiao, Songyang Gao, Chengqi Lyu, Yuzhe Gu, Wenwei Zhang, Derek~F Wong, Songyang Zhang, et~al.
\newblock Compassverifier: A unified and robust verifier for llms evaluation and outcome reward.
\newblock \emph{arXiv preprint arXiv:2508.03686}, 2025{\natexlab{a}}.

\bibitem[Liu et~al.(2023)Liu, Zeng, He, Jiang, and He]{liu2023makes}
Wei Liu, Weihao Zeng, Keqing He, Yong Jiang, and Junxian He.
\newblock What makes good data for alignment? a comprehensive study of automatic data selection in instruction tuning.
\newblock \emph{arXiv preprint arXiv:2312.15685}, 2023.

\bibitem[Liu et~al.(2025{\natexlab{b}})Liu, Chen, Li, Qi, Pang, Du, Lee, and Lin]{liu2025understanding}
Zichen Liu, Changyu Chen, Wenjun Li, Penghui Qi, Tianyu Pang, Chao Du, Wee~Sun Lee, and Min Lin.
\newblock Understanding r1-zero-like training: A critical perspective.
\newblock \emph{arXiv preprint arXiv:2503.20783}, 2025{\natexlab{b}}.

\bibitem[Liu et~al.(2025{\natexlab{c}})Liu, Meng, Du, Zhou, Yu, Shao, and Zhang]{liu2025cpgd}
Zongkai Liu, Fanqing Meng, Lingxiao Du, Zhixiang Zhou, Chao Yu, Wenqi Shao, and Qiaosheng Zhang.
\newblock Cpgd: Toward stable rule-based reinforcement learning for language models.
\newblock \emph{arXiv preprint arXiv:2505.12504}, 2025{\natexlab{c}}.

\bibitem[Lu et~al.(2023)Lu, Bansal, Xia, Liu, Li, Hajishirzi, Cheng, Chang, Galley, and Gao]{lu2023mathvista}
Pan Lu, Hritik Bansal, Tony Xia, Jiacheng Liu, Chunyuan Li, Hannaneh Hajishirzi, Hao Cheng, Kai-Wei Chang, Michel Galley, and Jianfeng Gao.
\newblock Mathvista: Evaluating mathematical reasoning of foundation models in visual contexts.
\newblock \emph{arXiv preprint arXiv:2310.02255}, 2023.

\bibitem[Luo et~al.(2025)Luo, Wen, Hu, Sun, Liu, Sun, Lyu, and Chen]{luo2025multi}
Kairong Luo, Haodong Wen, Shengding Hu, Zhenbo Sun, Zhiyuan Liu, Maosong Sun, Kaifeng Lyu, and Wenguang Chen.
\newblock A multi-power law for loss curve prediction across learning rate schedules.
\newblock \emph{arXiv preprint arXiv:2503.12811}, 2025.

\bibitem[Lyu et~al.(2025)Lyu, Gao, Gu, Zhang, Gao, Liu, Wang, Li, Zhao, Huang, et~al.]{lyu2025exploring}
Chengqi Lyu, Songyang Gao, Yuzhe Gu, Wenwei Zhang, Jianfei Gao, Kuikun Liu, Ziyi Wang, Shuaibin Li, Qian Zhao, Haian Huang, et~al.
\newblock Exploring the limit of outcome reward for learning mathematical reasoning.
\newblock \emph{arXiv preprint arXiv:2502.06781}, 2025.

\bibitem[McCandlish et~al.(2018)McCandlish, Kaplan, Amodei, and Team]{mccandlish2018empirical}
Sam McCandlish, Jared Kaplan, Dario Amodei, and OpenAI~Dota Team.
\newblock An empirical model of large-batch training.
\newblock \emph{arXiv preprint arXiv:1812.06162}, 2018.

\bibitem[Meng et~al.(2025)Meng, Du, Liu, Zhou, Lu, Fu, Shi, Wang, He, Zhang, et~al.]{meng2025mm}
Fanqing Meng, Lingxiao Du, Zongkai Liu, Zhixiang Zhou, Quanfeng Lu, Daocheng Fu, Botian Shi, Wenhai Wang, Junjun He, Kaipeng Zhang, et~al.
\newblock Mm-eureka: Exploring visual aha moment with rule-based large-scale reinforcement learning.
\newblock \emph{CoRR}, 2025.

\bibitem[Mirza et~al.(2025)Mirza, Alampara, Kunchapu, R{\'\i}os-Garc{\'\i}a, Emoekabu, Krishnan, Gupta, Schilling-Wilhelmi, Okereke, Aneesh, et~al.]{mirza2025framework}
Adrian Mirza, Nawaf Alampara, Sreekanth Kunchapu, Marti{\~n}o R{\'\i}os-Garc{\'\i}a, Benedict Emoekabu, Aswanth Krishnan, Tanya Gupta, Mara Schilling-Wilhelmi, Macjonathan Okereke, Anagha Aneesh, et~al.
\newblock A framework for evaluating the chemical knowledge and reasoning abilities of large language models against the expertise of chemists.
\newblock \emph{Nature Chemistry}, pp.\  1--8, 2025.

\bibitem[OpenAI(2025)]{o3}
OpenAI.
\newblock Introducing openai o3 and o4-mini, 2025.
\newblock URL \url{https://openai.com/index/introducing-o3-and-o4-mini/}.

\bibitem[OpenAI et~al.(2024)OpenAI, Achiam, Adler, Agarwal, Ahmad, Akkaya, Aleman, Almeida, Altenschmidt, Altman, Anadkat, Avila, Babuschkin, Balaji, Balcom, Baltescu, Bao, Bavarian, Belgum, Bello, Berdine, Bernadett-Shapiro, Berner, Bogdonoff, Boiko, Boyd, Brakman, Brockman, Brooks, Brundage, Button, Cai, Campbell, Cann, Carey, Carlson, Carmichael, Chan, Chang, Chantzis, Chen, Chen, Chen, Chen, Chen, Chess, Cho, Chu, Chung, Cummings, Currier, Dai, Decareaux, Degry, Deutsch, Deville, Dhar, Dohan, Dowling, Dunning, Ecoffet, Eleti, Eloundou, Farhi, Fedus, Felix, Fishman, Forte, Fulford, Gao, Georges, Gibson, Goel, Gogineni, Goh, Gontijo-Lopes, Gordon, Grafstein, Gray, Greene, Gross, Gu, Guo, Hallacy, Han, Harris, He, Heaton, Heidecke, Hesse, Hickey, Hickey, Hoeschele, Houghton, Hsu, Hu, Hu, Huizinga, Jain, Jain, Jang, Jiang, Jiang, Jin, Jin, Jomoto, Jonn, Jun, Kaftan, Łukasz Kaiser, Kamali, Kanitscheider, Keskar, Khan, Kilpatrick, Kim, Kim, Kim, Kirchner, Kiros, Knight, Kokotajlo, Łukasz Kondraciuk, Kondrich,
  Konstantinidis, Kosic, Krueger, Kuo, Lampe, Lan, Lee, Leike, Leung, Levy, Li, Lim, Lin, Lin, Litwin, Lopez, Lowe, Lue, Makanju, Malfacini, Manning, Markov, Markovski, Martin, Mayer, Mayne, McGrew, McKinney, McLeavey, McMillan, McNeil, Medina, Mehta, Menick, Metz, Mishchenko, Mishkin, Monaco, Morikawa, Mossing, Mu, Murati, Murk, Mély, Nair, Nakano, Nayak, Neelakantan, Ngo, Noh, Ouyang, O'Keefe, Pachocki, Paino, Palermo, Pantuliano, Parascandolo, Parish, Parparita, Passos, Pavlov, Peng, Perelman, de~Avila Belbute~Peres, Petrov, de~Oliveira~Pinto, Michael, Pokorny, Pokrass, Pong, Powell, Power, Power, Proehl, Puri, Radford, Rae, Ramesh, Raymond, Real, Rimbach, Ross, Rotsted, Roussez, Ryder, Saltarelli, Sanders, Santurkar, Sastry, Schmidt, Schnurr, Schulman, Selsam, Sheppard, Sherbakov, Shieh, Shoker, Shyam, Sidor, Sigler, Simens, Sitkin, Slama, Sohl, Sokolowsky, Song, Staudacher, Such, Summers, Sutskever, Tang, Tezak, Thompson, Tillet, Tootoonchian, Tseng, Tuggle, Turley, Tworek, Uribe, Vallone, Vijayvergiya,
  Voss, Wainwright, Wang, Wang, Wang, Ward, Wei, Weinmann, Welihinda, Welinder, Weng, Weng, Wiethoff, Willner, Winter, Wolrich, Wong, Workman, Wu, Wu, Wu, Xiao, Xu, Yoo, Yu, Yuan, Zaremba, Zellers, Zhang, Zhang, Zhao, Zheng, Zhuang, Zhuk, and Zoph]{openai2024gpt4technicalreport}
OpenAI, Josh Achiam, Steven Adler, Sandhini Agarwal, Lama Ahmad, Ilge Akkaya, Florencia~Leoni Aleman, Diogo Almeida, Janko Altenschmidt, Sam Altman, Shyamal Anadkat, Red Avila, Igor Babuschkin, Suchir Balaji, Valerie Balcom, Paul Baltescu, Haiming Bao, Mohammad Bavarian, Jeff Belgum, Irwan Bello, Jake Berdine, Gabriel Bernadett-Shapiro, Christopher Berner, Lenny Bogdonoff, Oleg Boiko, Madelaine Boyd, Anna-Luisa Brakman, Greg Brockman, Tim Brooks, Miles Brundage, Kevin Button, Trevor Cai, Rosie Campbell, Andrew Cann, Brittany Carey, Chelsea Carlson, Rory Carmichael, Brooke Chan, Che Chang, Fotis Chantzis, Derek Chen, Sully Chen, Ruby Chen, Jason Chen, Mark Chen, Ben Chess, Chester Cho, Casey Chu, Hyung~Won Chung, Dave Cummings, Jeremiah Currier, Yunxing Dai, Cory Decareaux, Thomas Degry, Noah Deutsch, Damien Deville, Arka Dhar, David Dohan, Steve Dowling, Sheila Dunning, Adrien Ecoffet, Atty Eleti, Tyna Eloundou, David Farhi, Liam Fedus, Niko Felix, Simón~Posada Fishman, Juston Forte, Isabella Fulford, Leo
  Gao, Elie Georges, Christian Gibson, Vik Goel, Tarun Gogineni, Gabriel Goh, Rapha Gontijo-Lopes, Jonathan Gordon, Morgan Grafstein, Scott Gray, Ryan Greene, Joshua Gross, Shixiang~Shane Gu, Yufei Guo, Chris Hallacy, Jesse Han, Jeff Harris, Yuchen He, Mike Heaton, Johannes Heidecke, Chris Hesse, Alan Hickey, Wade Hickey, Peter Hoeschele, Brandon Houghton, Kenny Hsu, Shengli Hu, Xin Hu, Joost Huizinga, Shantanu Jain, Shawn Jain, Joanne Jang, Angela Jiang, Roger Jiang, Haozhun Jin, Denny Jin, Shino Jomoto, Billie Jonn, Heewoo Jun, Tomer Kaftan, Łukasz Kaiser, Ali Kamali, Ingmar Kanitscheider, Nitish~Shirish Keskar, Tabarak Khan, Logan Kilpatrick, Jong~Wook Kim, Christina Kim, Yongjik Kim, Jan~Hendrik Kirchner, Jamie Kiros, Matt Knight, Daniel Kokotajlo, Łukasz Kondraciuk, Andrew Kondrich, Aris Konstantinidis, Kyle Kosic, Gretchen Krueger, Vishal Kuo, Michael Lampe, Ikai Lan, Teddy Lee, Jan Leike, Jade Leung, Daniel Levy, Chak~Ming Li, Rachel Lim, Molly Lin, Stephanie Lin, Mateusz Litwin, Theresa Lopez, Ryan
  Lowe, Patricia Lue, Anna Makanju, Kim Malfacini, Sam Manning, Todor Markov, Yaniv Markovski, Bianca Martin, Katie Mayer, Andrew Mayne, Bob McGrew, Scott~Mayer McKinney, Christine McLeavey, Paul McMillan, Jake McNeil, David Medina, Aalok Mehta, Jacob Menick, Luke Metz, Andrey Mishchenko, Pamela Mishkin, Vinnie Monaco, Evan Morikawa, Daniel Mossing, Tong Mu, Mira Murati, Oleg Murk, David Mély, Ashvin Nair, Reiichiro Nakano, Rajeev Nayak, Arvind Neelakantan, Richard Ngo, Hyeonwoo Noh, Long Ouyang, Cullen O'Keefe, Jakub Pachocki, Alex Paino, Joe Palermo, Ashley Pantuliano, Giambattista Parascandolo, Joel Parish, Emy Parparita, Alex Passos, Mikhail Pavlov, Andrew Peng, Adam Perelman, Filipe de~Avila Belbute~Peres, Michael Petrov, Henrique~Ponde de~Oliveira~Pinto, Michael, Pokorny, Michelle Pokrass, Vitchyr~H. Pong, Tolly Powell, Alethea Power, Boris Power, Elizabeth Proehl, Raul Puri, Alec Radford, Jack Rae, Aditya Ramesh, Cameron Raymond, Francis Real, Kendra Rimbach, Carl Ross, Bob Rotsted, Henri Roussez,
  Nick Ryder, Mario Saltarelli, Ted Sanders, Shibani Santurkar, Girish Sastry, Heather Schmidt, David Schnurr, John Schulman, Daniel Selsam, Kyla Sheppard, Toki Sherbakov, Jessica Shieh, Sarah Shoker, Pranav Shyam, Szymon Sidor, Eric Sigler, Maddie Simens, Jordan Sitkin, Katarina Slama, Ian Sohl, Benjamin Sokolowsky, Yang Song, Natalie Staudacher, Felipe~Petroski Such, Natalie Summers, Ilya Sutskever, Jie Tang, Nikolas Tezak, Madeleine~B. Thompson, Phil Tillet, Amin Tootoonchian, Elizabeth Tseng, Preston Tuggle, Nick Turley, Jerry Tworek, Juan Felipe~Cerón Uribe, Andrea Vallone, Arun Vijayvergiya, Chelsea Voss, Carroll Wainwright, Justin~Jay Wang, Alvin Wang, Ben Wang, Jonathan Ward, Jason Wei, CJ~Weinmann, Akila Welihinda, Peter Welinder, Jiayi Weng, Lilian Weng, Matt Wiethoff, Dave Willner, Clemens Winter, Samuel Wolrich, Hannah Wong, Lauren Workman, Sherwin Wu, Jeff Wu, Michael Wu, Kai Xiao, Tao Xu, Sarah Yoo, Kevin Yu, Qiming Yuan, Wojciech Zaremba, Rowan Zellers, Chong Zhang, Marvin Zhang, Shengjia
  Zhao, Tianhao Zheng, Juntang Zhuang, William Zhuk, and Barret Zoph.
\newblock Gpt-4 technical report, 2024.
\newblock URL \url{https://arxiv.org/abs/2303.08774}.

\bibitem[{PrismaX}(2025)]{msearthmcq2025hf}
{PrismaX}.
\newblock Msearth\_mcq.
\newblock \url{https://huggingface.co/datasets/PrismaX/MSEarth_MCQ}, 2025.
\newblock Version with 2.78K multiple-choice items; accessed 2025-08-18.

\bibitem[Provilkov et~al.(2019)Provilkov, Emelianenko, and Voita]{provilkov2019bpe}
Ivan Provilkov, Dmitrii Emelianenko, and Elena Voita.
\newblock Bpe-dropout: Simple and effective subword regularization.
\newblock \emph{arXiv preprint arXiv:1910.13267}, 2019.

\bibitem[Rein et~al.(2024)Rein, Hou, Stickland, Petty, Pang, Dirani, Michael, and Bowman]{rein2024gpqa}
David Rein, Betty~Li Hou, Asa~Cooper Stickland, Jackson Petty, Richard~Yuanzhe Pang, Julien Dirani, Julian Michael, and Samuel~R Bowman.
\newblock Gpqa: A graduate-level google-proof q\&a benchmark.
\newblock In \emph{First Conference on Language Modeling}, 2024.

\bibitem[Shao et~al.(2024)Shao, Wang, Zhu, Xu, Song, Bi, Zhang, Zhang, Li, Wu, et~al.]{shao2024deepseekmath}
Zhihong Shao, Peiyi Wang, Qihao Zhu, Runxin Xu, Junxiao Song, Xiao Bi, Haowei Zhang, Mingchuan Zhang, YK~Li, Yang Wu, et~al.
\newblock Deepseekmath: Pushing the limits of mathematical reasoning in open language models.
\newblock \emph{arXiv preprint arXiv:2402.03300}, 2024.

\bibitem[Shen et~al.(2025)Shen, Zhao, Gu, Gao, Liu, Huang, Gao, Lin, Zhang, and Chen]{shen2025semioffpolicyreinforcementlearningvisionlanguage}
Junhao Shen, Haiteng Zhao, Yuzhe Gu, Songyang Gao, Kuikun Liu, Haian Huang, Jianfei Gao, Dahua Lin, Wenwei Zhang, and Kai Chen.
\newblock Semi-off-policy reinforcement learning for vision-language slow-thinking reasoning, 2025.
\newblock URL \url{https://arxiv.org/abs/2507.16814}.

\bibitem[Shen(2024)]{shen2024rethinking}
Ming Shen.
\newblock Rethinking data selection for supervised fine-tuning.
\newblock \emph{arXiv preprint arXiv:2402.06094}, 2024.

\bibitem[Shen et~al.(2024)Shen, Chen, Mamalakis, He, Xia, Li, Su, He, and Wang]{shen2024fine}
Yiqing Shen, Zan Chen, Michail Mamalakis, Luhan He, Haiyang Xia, Tianbin Li, Yanzhou Su, Junjun He, and Yu~Guang Wang.
\newblock A fine-tuning dataset and benchmark for large language models for protein understanding.
\newblock In \emph{2024 IEEE International Conference on Bioinformatics and Biomedicine (BIBM)}, pp.\  2390--2395. IEEE, 2024.

\bibitem[Sheng et~al.(2025)Sheng, Zhang, Ye, Wu, Zhang, Zhang, Peng, Lin, and Wu]{sheng2025hybridflow}
Guangming Sheng, Chi Zhang, Zilingfeng Ye, Xibin Wu, Wang Zhang, Ru~Zhang, Yanghua Peng, Haibin Lin, and Chuan Wu.
\newblock Hybridflow: A flexible and efficient rlhf framework.
\newblock In \emph{Proceedings of the Twentieth European Conference on Computer Systems}, pp.\  1279--1297, 2025.

\bibitem[Su et~al.(2024)Su, Kong, Lin, Jennings, Norick, Kliegl, Patwary, Shoeybi, and Catanzaro]{su2024nemotron}
Dan Su, Kezhi Kong, Ying Lin, Joseph Jennings, Brandon Norick, Markus Kliegl, Mostofa Patwary, Mohammad Shoeybi, and Bryan Catanzaro.
\newblock Nemotron-cc: Transforming common crawl into a refined long-horizon pretraining dataset.
\newblock \emph{arXiv preprint arXiv:2412.02595}, 2024.

\bibitem[Su et~al.(2025)Su, Fu, Cao, Gao, Jin, PeiZhilin, and Wang]{suzhongling2025tmaadaptive}
Zhongling Su, Rong Fu, Weihan Cao, Jianfei Gao, Minxi Jin, PeiZhilin, and Hui Wang.
\newblock {TMA}-adaptive {FP}8 grouped {GEMM}: Eliminating padding requirements in low-precision training and inference on hopper.
\newblock In \emph{ES-FoMo III: 3rd Workshop on Efficient Systems for Foundation Models}, 2025.
\newblock URL \url{https://openreview.net/forum?id=2upxraaj8H}.

\bibitem[Tang et~al.(2024)Tang, Ranjan, Pangarkar, Liang, Wang, An, Rao, Jin, Wang, Cheng, et~al.]{tang2024txt360}
Liping Tang, Nikhil Ranjan, Omkar Pangarkar, Xuezhi Liang, Zhen Wang, Li~An, Bhaskar Rao, Linghao Jin, Huijuan Wang, Zhoujun Cheng, et~al.
\newblock Txt360: A top-quality llm pre-training dataset requires the perfect blend, 2024.

\bibitem[Team et~al.(2025)Team, Du, Gao, Xing, Jiang, Chen, Li, Xiao, Du, Liao, et~al.]{team2025kimi}
Kimi Team, Angang Du, Bofei Gao, Bowei Xing, Changjiu Jiang, Cheng Chen, Cheng Li, Chenjun Xiao, Chenzhuang Du, Chonghua Liao, et~al.
\newblock Kimi k1. 5: Scaling reinforcement learning with llms.
\newblock \emph{arXiv preprint arXiv:2501.12599}, 2025.

\bibitem[Team(2024)]{team2024qwen2}
Qwen Team.
\newblock Qwen2 technical report.
\newblock \emph{arXiv preprint arXiv:2407.10671}, 2024.

\bibitem[Tissue et~al.(2024)Tissue, Wang, and Wang]{tissue2024scaling}
Howe Tissue, Venus Wang, and Lu~Wang.
\newblock Scaling law with learning rate annealing.
\newblock \emph{arXiv preprint arXiv:2408.11029}, 2024.

\bibitem[Wang et~al.(2024{\natexlab{a}})Wang, Xu, Zhao, Ouyang, Wu, Zhao, Xu, Liu, Qu, Shang, Zhang, Wei, Sui, Li, Shi, Qiao, Lin, and He]{wang2024mineruopensourcesolutionprecise}
Bin Wang, Chao Xu, Xiaomeng Zhao, Linke Ouyang, Fan Wu, Zhiyuan Zhao, Rui Xu, Kaiwen Liu, Yuan Qu, Fukai Shang, Bo~Zhang, Liqun Wei, Zhihao Sui, Wei Li, Botian Shi, Yu~Qiao, Dahua Lin, and Conghui He.
\newblock Mineru: An open-source solution for precise document content extraction, 2024{\natexlab{a}}.
\newblock URL \url{https://arxiv.org/abs/2409.18839}.

\bibitem[Wang et~al.(2025{\natexlab{a}})Wang, Wang, Chen, Wang, Wang, Guo, Ma, Lan, Yang, Zhang, Liu, and Sun]{wang2025xlrsbenchmultimodalllmsunderstand}
Fengxiang Wang, Hongzhen Wang, Mingshuo Chen, Di~Wang, Yulin Wang, Zonghao Guo, Qiang Ma, Long Lan, Wenjing Yang, Jing Zhang, Zhiyuan Liu, and Maosong Sun.
\newblock Xlrs-bench: Could your multimodal llms understand extremely large ultra-high-resolution remote sensing imagery?, 2025{\natexlab{a}}.
\newblock URL \url{https://arxiv.org/abs/2503.23771}.

\bibitem[Wang et~al.(2024{\natexlab{b}})Wang, Pan, Shi, Lu, Ren, Zhou, Zhan, and Li]{wang2024measuring}
Ke~Wang, Junting Pan, Weikang Shi, Zimu Lu, Houxing Ren, Aojun Zhou, Mingjie Zhan, and Hongsheng Li.
\newblock Measuring multimodal mathematical reasoning with math-vision dataset.
\newblock \emph{Advances in Neural Information Processing Systems}, 37:\penalty0 95095--95169, 2024{\natexlab{b}}.

\bibitem[Wang et~al.(2025{\natexlab{b}})Wang, Yu, Gao, Zheng, Liu, Lu, Dang, Chen, Yang, Zhang, et~al.]{wang2025beyond}
Shenzhi Wang, Le~Yu, Chang Gao, Chujie Zheng, Shixuan Liu, Rui Lu, Kai Dang, Xionghui Chen, Jianxin Yang, Zhenru Zhang, et~al.
\newblock Beyond the 80/20 rule: High-entropy minority tokens drive effective reinforcement learning for llm reasoning.
\newblock \emph{arXiv preprint arXiv:2506.01939}, 2025{\natexlab{b}}.

\bibitem[Wang et~al.(2024{\natexlab{c}})Wang, Chen, Wang, Cao, Liu, Gao, Zhu, Zhu, Lu, Qiao, et~al.]{wang2024enhancing}
Weiyun Wang, Zhe Chen, Wenhai Wang, Yue Cao, Yangzhou Liu, Zhangwei Gao, Jinguo Zhu, Xizhou Zhu, Lewei Lu, Yu~Qiao, et~al.
\newblock Enhancing the reasoning ability of multimodal large language models via mixed preference optimization.
\newblock \emph{arXiv preprint arXiv:2411.10442}, 2024{\natexlab{c}}.

\bibitem[Wang et~al.(2024{\natexlab{d}})Wang, Ma, Zhang, Ni, Chandra, Guo, Ren, Arulraj, He, Jiang, et~al.]{wang2024mmlu}
Yubo Wang, Xueguang Ma, Ge~Zhang, Yuansheng Ni, Abhranil Chandra, Shiguang Guo, Weiming Ren, Aaran Arulraj, Xuan He, Ziyan Jiang, et~al.
\newblock Mmlu-pro: A more robust and challenging multi-task language understanding benchmark.
\newblock \emph{Advances in Neural Information Processing Systems}, 37:\penalty0 95266--95290, 2024{\natexlab{d}}.

\bibitem[Ward et~al.(2025)Ward, Lin, Venhoff, and Nanda]{ward2025reasoning}
Jake Ward, Chuqiao Lin, Constantin Venhoff, and Neel Nanda.
\newblock Reasoning-finetuning repurposes latent representations in base models.
\newblock \emph{arXiv preprint arXiv:2507.12638}, 2025.

\bibitem[xAI(2025)]{grok4}
xAI.
\newblock Grok 4, 2025.
\newblock URL \url{https://x.ai/news/grok-4}.

\bibitem[Xia et~al.(2025)Xia, Jin, Xie, He, Cao, Luo, Liu, Wang, Liu, Chen, et~al.]{xia2025naturelm}
Yingce Xia, Peiran Jin, Shufang Xie, Liang He, Chuan Cao, Renqian Luo, Guoqing Liu, Yue Wang, Zequn Liu, Yuan-Jyue Chen, et~al.
\newblock Naturelm: Deciphering the language of nature for scientific discovery.
\newblock \emph{arXiv e-prints}, pp.\  arXiv--2502, 2025.

\bibitem[Xiaomi et~al.(2025)Xiaomi, Xia, Shen, Zhu, Zhang, Wang, Zhang, Liu, Xiao, Dong, et~al.]{xiaomi2025mimo}
LLM Xiaomi, Bingquan Xia, Bowen Shen, Dawei Zhu, Di~Zhang, Gang Wang, Hailin Zhang, Huaqiu Liu, Jiebao Xiao, Jinhao Dong, et~al.
\newblock Mimo: Unlocking the reasoning potential of language model--from pretraining to posttraining.
\newblock \emph{arXiv preprint arXiv:2505.07608}, 2025.

\bibitem[Yang et~al.(2025)Yang, Li, Yang, Zhang, Hui, Zheng, Yu, Gao, Huang, Lv, et~al.]{yang2025qwen3}
An~Yang, Anfeng Li, Baosong Yang, Beichen Zhang, Binyuan Hui, Bo~Zheng, Bowen Yu, Chang Gao, Chengen Huang, Chenxu Lv, et~al.
\newblock Qwen3 technical report.
\newblock \emph{arXiv preprint arXiv:2505.09388}, 2025.

\bibitem[Yang et~al.(2021)Yang, Hu, Babuschkin, Sidor, Liu, Farhi, Ryder, Pachocki, Chen, and Gao]{yang2021tuning}
Ge~Yang, Edward Hu, Igor Babuschkin, Szymon Sidor, Xiaodong Liu, David Farhi, Nick Ryder, Jakub Pachocki, Weizhu Chen, and Jianfeng Gao.
\newblock Tuning large neural networks via zero-shot hyperparameter transfer.
\newblock \emph{Advances in Neural Information Processing Systems}, 34:\penalty0 17084--17097, 2021.

\bibitem[Yu et~al.(2024)Yu, Baker, Chen, Ning, and Sun]{yu2024llasmol}
Botao Yu, Frazier~N. Baker, Ziqi Chen, Xia Ning, and Huan Sun.
\newblock Llasmol: Advancing large language models for chemistry with a large-scale, comprehensive, high-quality instruction tuning dataset.
\newblock \emph{arXiv preprint arXiv:2402.09391}, 2024.

\bibitem[Yu et~al.(2025)Yu, Zhang, Zhu, Yuan, Zuo, Yue, Dai, Fan, Liu, Liu, et~al.]{yu2025dapo}
Qiying Yu, Zheng Zhang, Ruofei Zhu, Yufeng Yuan, Xiaochen Zuo, Yu~Yue, Weinan Dai, Tiantian Fan, Gaohong Liu, Lingjun Liu, et~al.
\newblock Dapo: An open-source llm reinforcement learning system at scale.
\newblock \emph{arXiv preprint arXiv:2503.14476}, 2025.

\bibitem[Yue et~al.(2024)Yue, Ni, Zhang, Zheng, Liu, Zhang, Stevens, Jiang, Ren, Sun, et~al.]{yue2024mmmu}
Xiang Yue, Yuansheng Ni, Kai Zhang, Tianyu Zheng, Ruoqi Liu, Ge~Zhang, Samuel Stevens, Dongfu Jiang, Weiming Ren, Yuxuan Sun, et~al.
\newblock Mmmu: A massive multi-discipline multimodal understanding and reasoning benchmark for expert agi.
\newblock In \emph{Proceedings of the IEEE/CVF Conference on Computer Vision and Pattern Recognition}, pp.\  9556--9567, 2024.

\bibitem[Zhao et~al.(2025)Zhao, Xu, Liu, Zhou, Ling, Fei, Yue, Bai, Zhang, and Wu]{zhao2025msearth}
Xiangyu Zhao, Wanghan Xu, Bo~Liu, Yuhao Zhou, Fenghua Ling, Ben Fei, Xiaoyu Yue, Lei Bai, Wenlong Zhang, and Xiao-Ming Wu.
\newblock Msearth: A benchmark for multimodal scientific comprehension of earth science.
\newblock \emph{arXiv preprint arXiv:2505.20740}, 2025.

\bibitem[Zheng et~al.(2025)Zheng, Liu, Li, Chen, Yu, Gao, Dang, Liu, Men, Yang, Zhou, and Lin]{zheng2025group}
Chujie Zheng, Shixuan Liu, Mingze Li, Xiong-Hui Chen, Bowen Yu, Chang Gao, Kai Dang, Yuqiong Liu, Rui Men, An~Yang, Jingren Zhou, and Junyang Lin.
\newblock Group sequence policy optimization.
\newblock \emph{arXiv preprint arXiv:2507.18071}, 2025.

\bibitem[Zhou et~al.(2023)Zhou, Lu, Mishra, Brahma, Basu, Luan, Zhou, and Hou]{zhou2023instruction}
Jeffrey Zhou, Tianjian Lu, Swaroop Mishra, Siddhartha Brahma, Sujoy Basu, Yi~Luan, Denny Zhou, and Le~Hou.
\newblock Instruction-following evaluation for large language models.
\newblock \emph{arXiv preprint arXiv:2311.07911}, 2023.

\bibitem[Zhou et~al.(2025)Zhou, Wang, He, Xiao, Li, Feng, Guo, Yang, Wu, Huang, et~al.]{zhou2025scientists}
Yuhao Zhou, Yiheng Wang, Xuming He, Ruoyao Xiao, Zhiwei Li, Qiantai Feng, Zijie Guo, Yuejin Yang, Hao Wu, Wenxuan Huang, et~al.
\newblock Scientists' first exam: Probing cognitive abilities of mllm via perception, understanding, and reasoning.
\newblock \emph{arXiv preprint arXiv:2506.10521}, 2025.

\bibitem[Zhu et~al.(2025)Zhu, Wang, Chen, Liu, Ye, Gu, Tian, Duan, Su, Shao, et~al.]{zhu2025internvl3}
Jinguo Zhu, Weiyun Wang, Zhe Chen, Zhaoyang Liu, Shenglong Ye, Lixin Gu, Hao Tian, Yuchen Duan, Weijie Su, Jie Shao, et~al.
\newblock Internvl3: Exploring advanced training and test-time recipes for open-source multimodal models.
\newblock \emph{arXiv preprint arXiv:2504.10479}, 2025.

\end{thebibliography}
\bibliographystyle{iclr2025_conference}

\end{document}